\documentclass[11pt, a4paper, copyright]{google}

\usepackage{tikz}
\usepackage{multicol}
\usetikzlibrary{positioning,arrows.meta,shapes.geometric,fit,backgrounds,calc,decorations.pathreplacing}

\usepackage[authoryear, sort&compress, round]{natbib}
\makeatletter
\renewcommand{\absfont}{\normalfont\linespread{1.2}\fontsize{11}{12}\selectfont}

\definecolor{princetonorange}{HTML}{E77500}
\definecolor{absboxbg}{HTML}{FFF8F0}
\definecolor{absboxframe}{HTML}{E77500}

\makeatother

\definecolor{linkblue}{RGB}{0,0,139}
\definecolor{navy}{RGB}{0,0,128}
\definecolor{royalblue}{RGB}{65,105,225}
\definecolor{steelblue}{RGB}{70,130,180}
\definecolor{dodgerblue}{RGB}{30,144,255}
\definecolor{mediumblue}{RGB}{0,0,205}
\definecolor{darkslateblue}{RGB}{72,61,139}
\usepackage[colorlinks = true,
            linkcolor = linkblue,
            urlcolor  = royalblue,
            citecolor = brown,
            anchorcolor = blue]{hyperref}

\usepackage[misc]{ifsym}
\usepackage{minitoc}
\usepackage{cleveref}     
\usepackage{subcaption}
\usepackage{booktabs}
\usepackage{amsmath}
\usepackage{mathtools}
\usepackage{amssymb}
\usepackage{graphicx}
\usepackage{algorithmic}
\usepackage{algorithm}
\usepackage{tcolorbox}
\usepackage{verbatim}
\usepackage{makecell}
\usepackage{array}
\usepackage{multirow}
\usepackage{xurl}
\usepackage{amsthm}
\usepackage{adjustbox}
\usepackage{caption}
\usepackage{xcolor}
\usepackage{wrapfig}
\tcbuselibrary{skins,breakable}
\definecolor{caseink}{RGB}{38,48,66}
\definecolor{casetint}{RGB}{238,242,248}
\definecolor{codetint}{RGB}{246,246,243}
\definecolor{coderule}{RGB}{201,203,207}
\definecolor{caseaccent}{RGB}{192,62,38}
\newtcolorbox{casepanel}[1]{%
  breakable, enhanced, colback=white, colframe=caseink,
  boxrule=0.7pt, arc=1.8pt, left=3mm, right=3mm, top=2.6mm, bottom=2.6mm,
  title={#1}, coltitle=caseink, colbacktitle=casetint,
  fonttitle=\bfseries\small,
  attach boxed title to top left={xshift=4mm, yshift=-2.4mm},
  boxed title style={colframe=caseink, boxrule=0.6pt, arc=1.2pt},
  }
\newtcolorbox{artifactbox}[1]{%
  enhanced, colback=codetint, colframe=coderule, boxrule=0.5pt, arc=1pt,
  left=2.5mm, right=2.5mm, top=1.5mm, bottom=1.5mm,
  fontupper=\ttfamily\scriptsize, before skip=2mm, after skip=2mm,
  overlay={\node[anchor=north east, font=\sffamily\tiny, text=black!50,
    inner sep=2.5pt] at (frame.north east) {#1};}}

\usepackage{float}
\usepackage{placeins}   

\usepackage{pifont}
\usepackage{enumitem}

\usepackage[titles]{tocloft}
\definecolor{tocsubsec}{HTML}{444444}
\definecolor{tocsubsubsec}{HTML}{666666}

\usepackage{soul}

\usepackage{listings}
\tcbuselibrary{listings,skins,breakable}

\lstdefinestyle{prompt}{
  basicstyle=\ttfamily\footnotesize,
  breaklines=true,
  breakatwhitespace=true,
  columns=fullflexible,
  keepspaces=true,
  showstringspaces=false,
  postbreak=\mbox{\textcolor{gray}{$\hookrightarrow$}\space}
}

\usepackage{xspace}
\newcommand{\ours}{Recuris\xspace}

\usepackage{colortbl}     
\newcommand{\best}[1]{\textbf{#1}}
\newcommand{\sbest}[1]{\underline{#1}}
\definecolor{gainlo}{RGB}{243,109,0}
\definecolor{gainhi}{RGB}{206,0,20}
\newcommand{\heatmax}{20}
\newcommand{\heatgamma}{0.62}   
\newcommand{\hc}[2]{%
  \begingroup
  \edef\heatpct{\fpeval{round(100*(min(1,max(0,#1)/\heatmax))**\heatgamma,0)}}%
  \colorlet{heatmix}{gainhi!\heatpct!gainlo}%
  \textcolor{heatmix}{#2}%
  \endgroup}
\newcommand{\gain}[1]{\hc{#1}{\footnotesize\bfseries\,($+$#1)}}
\newcommand{\gaindag}[1]{\hc{#1}{\footnotesize\bfseries\,($+$#1$^{\dagger}$)}}
\newcommand{\drop}[1]{\textcolor{black!55}{\scriptsize\,($-$#1)}}
\newcommand{\oursrow}{\rowcolor[gray]{0.90}}
\newcommand{\na}{\textendash}                    
\newcommand{\phgain}[1]{\hc{#1}{\footnotesize\bfseries\,($+$#1)}}

\definecolor{accent1}{HTML}{2E6DA4}
\definecolor{accent2}{HTML}{D9534F}
\definecolor{accent3}{HTML}{5CB85C}
\definecolor{lightblue}{RGB}{173,216,230}
\definecolor{lightorange}{RGB}{255,213,170}
\definecolor{lightgreen}{RGB}{176,226,176}
\definecolor{lightyellow}{RGB}{255,255,204}
\definecolor{lightgray}{RGB}{220,220,220}
\definecolor{lightpurple}{RGB}{221,160,221}
\definecolor{lightred}{RGB}{255,182,193}
\definecolor{gray60}{gray}{0.6}
\definecolor{accent4}{HTML}{9B59B6}

\makeatletter
\renewcommand{\abscontent}{%
  \begin{center}
  {\fontsize{11pt}{13pt}\selectfont
   \href{https://github.com/Gen-Verse/Recuris}{\texttt{https://github.com/Gen-Verse/Recuris}}}
  \end{center}
  \vskip0.8em
  \noindent
  \parbox{\dimexpr\linewidth}{\absfont \theabstract}%
  \@ifundefined{@keywords}{}{%
    \vskip1em \noindent \keywordsfont Keywords: \@keywords}%
}
\renewcommand{\maketitle}{\bgroup\setlength{\parindent}{0pt}
  \begin{adjustwidth}{0pt}{24pt}
    \begin{center}
      {\titlefont \@title\par}%
      \vskip11pt
      {\@author\par}%
      \vskip20pt%
    \end{center}
  \end{adjustwidth}
  \egroup
  {\abscontent}%
  \thispagestyle{firststyle}
}
\makeatother

\newcommand{\hpara}[1]{\par\addvspace{0.7ex}\noindent\textbf{#1}\par\nobreak\noindent\ignorespaces}
\newcommand{\logoheight}{28pt}   
\AtBeginDocument{%
  \fancypagestyle{firststyle}{%
    \fancyhead[L]{}%
    \fancyhead[R]{}%
    \fancyhead[C]{%
      \IfFileExists{assets/recuris-logo.png}{%
        \includegraphics[height=\logoheight]{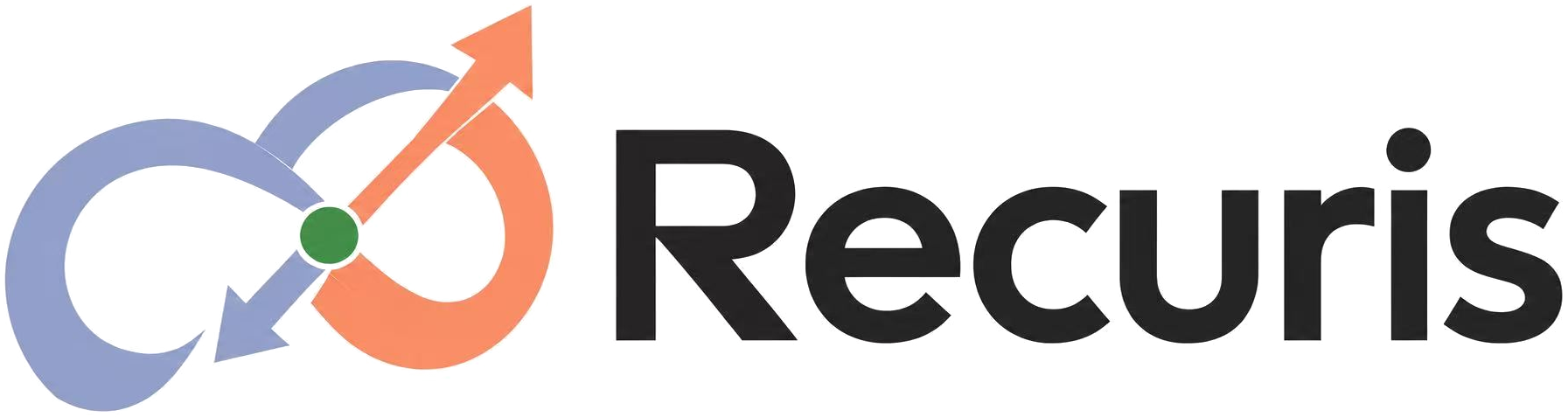}%
      }{}%
    }%
    \fancyfoot[L]{%
  $^{\dagger}$Corresponding authors;\quad
  \ifdefined\correspondingauthor
    \if\relax\the\correspondingauthor\relax
    \else
      {\itshape \the\correspondingauthor}%
    \fi
  \fi
  \\
  \ifthenelse{\boolean{copyright}}{\copyrightext}{}%
}%
    \fancyfoot[R]{\footerfont\bfseries\relax}%
    \fancyfoot[C]{\footerfont\bfseries\relax}%
  }%
}

\title{Recursive Experiential–Working Memory Evolution for Long-Horizon Agent Harnesses}

\correspondingauthor{Main contact: yangling0818@163.com}

 \makeatletter
 \renewcommand\AB@affilnote[1]{}
 \makeatother

\author{
\textbf{Zhaochen Yu}\textsuperscript{1,4} \quad
\textbf{Yingcheng Wu}\textsuperscript{2} \quad
\textbf{Zhenfei Yin}\textsuperscript{3} \quad
\textbf{Kaiyuan Chen} \quad
\textbf{Zhe Zhao}\textsuperscript{2}
\\
\textbf{Mengdi Wang}\textsuperscript{4,\(\dagger\)} \quad
\textbf{Shuicheng Yan}\textsuperscript{1,\(\dagger\)} \quad
\textbf{Ling Yang}\textsuperscript{4,\(\dagger\)}
\\
\normalfont\normalsize
\textsuperscript{1}NUS \quad
\textsuperscript{2}Stanford University \quad
\textsuperscript{3}University of Oxford \quad
\textsuperscript{4}Princeton University
}

\begin{abstract}

Recursive self-improvement (RSI) remains hard in long-horizon tasks, where growing histories obscure the task state and misalign skill invocation. We introduce \textbf{Recuris, a recursive Experiential--Working Memory architecture for long-horizon agent harnesses}, in which Working Memory tracks task progress and guides skill selection from Experiential Memory, grounding skill use in current needs rather than the full history. This coupling also turns execution into structured evidence that localizes failures to specific memory components. Across tasks, a fixed Meta-Agent turns that evidence into localized, validation-gated updates to Skill Memory that reshape execution and yield new evidence, forming a \textbf{bounded recursive memory-evolution loop}. Across four long-horizon benchmarks and ten models, \ours{} improves task success in 35 of the 37 completed model--benchmark pairs, carrying \textbf{frontier models to SOTA-level task success}: on $\tau^2$-Bench it adds $\mathbf{+17.8}$ points to \textbf{GPT-5.6 Sol} and $\mathbf{+15.6}$ to \textbf{Claude Opus 5}, taking Opus 5 to $\mathbf{87.9\%}$, and $+16.6$/$+13.5$ points on Qwen3.6-27B/35B on SkillFlow. The advantage widens as the interaction horizon grows, to $+32.2$ points on the longest tasks, and common long-horizon failures fall by up to $80\%$. These results position recursively evolving memory as a scalable foundation for RSI, enabling agents to continuously transform accumulated experience into increasingly effective long-horizon behavior.

\end{abstract}

\begin{document}

\maketitle

 \vspace{0.8ex}
\noindent\begin{minipage}{\linewidth}
  \centering
    \includegraphics[width=\linewidth]{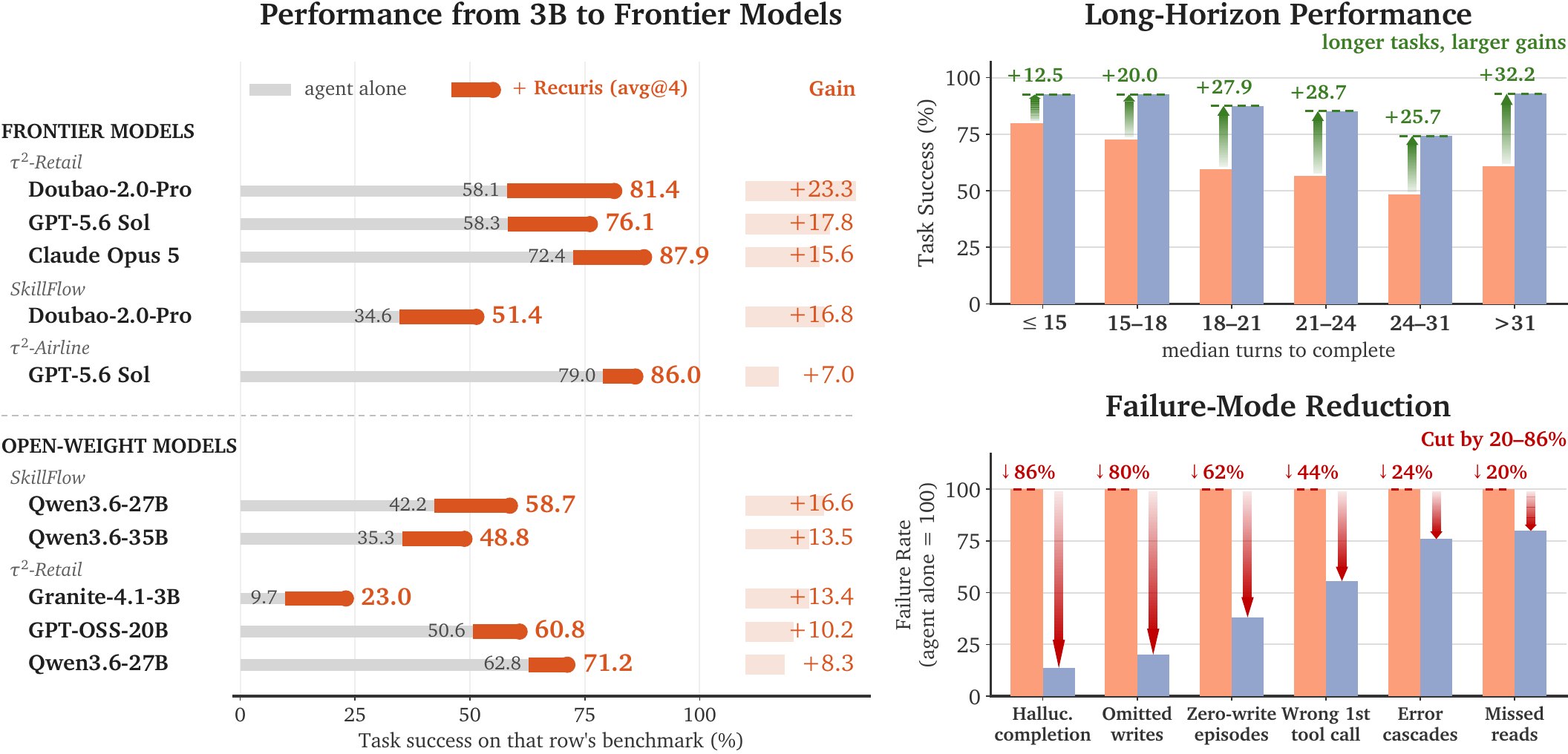}
  \vspace{-2.6ex}
  \captionof{figure}{%
    \textbf{\ours{} improves task success from 3B open-weight models to frontier
    models.} Performance across (i) task success (\%), (ii) task success by episode
    length, and (iii) six long-horizon failure modes, each rescaled so the
    agent alone $=100$, where a bar at 20 is an $80\%$ reduction.
  }
  \label{fig:hero}
\end{minipage}

\newpage
\vspace{0.5em}
{
  \hypersetup{linkcolor=black}
  \setlength{\parskip}{0pt}
  \renewcommand{\contentsname}{\normalfont\large\bfseries Contents}
  \setcounter{tocdepth}{3}
  \begingroup
    \small
    \tableofcontents
  \endgroup
}

\newpage

\begin{figure}[t]  \centering
  \includegraphics[width=\linewidth]{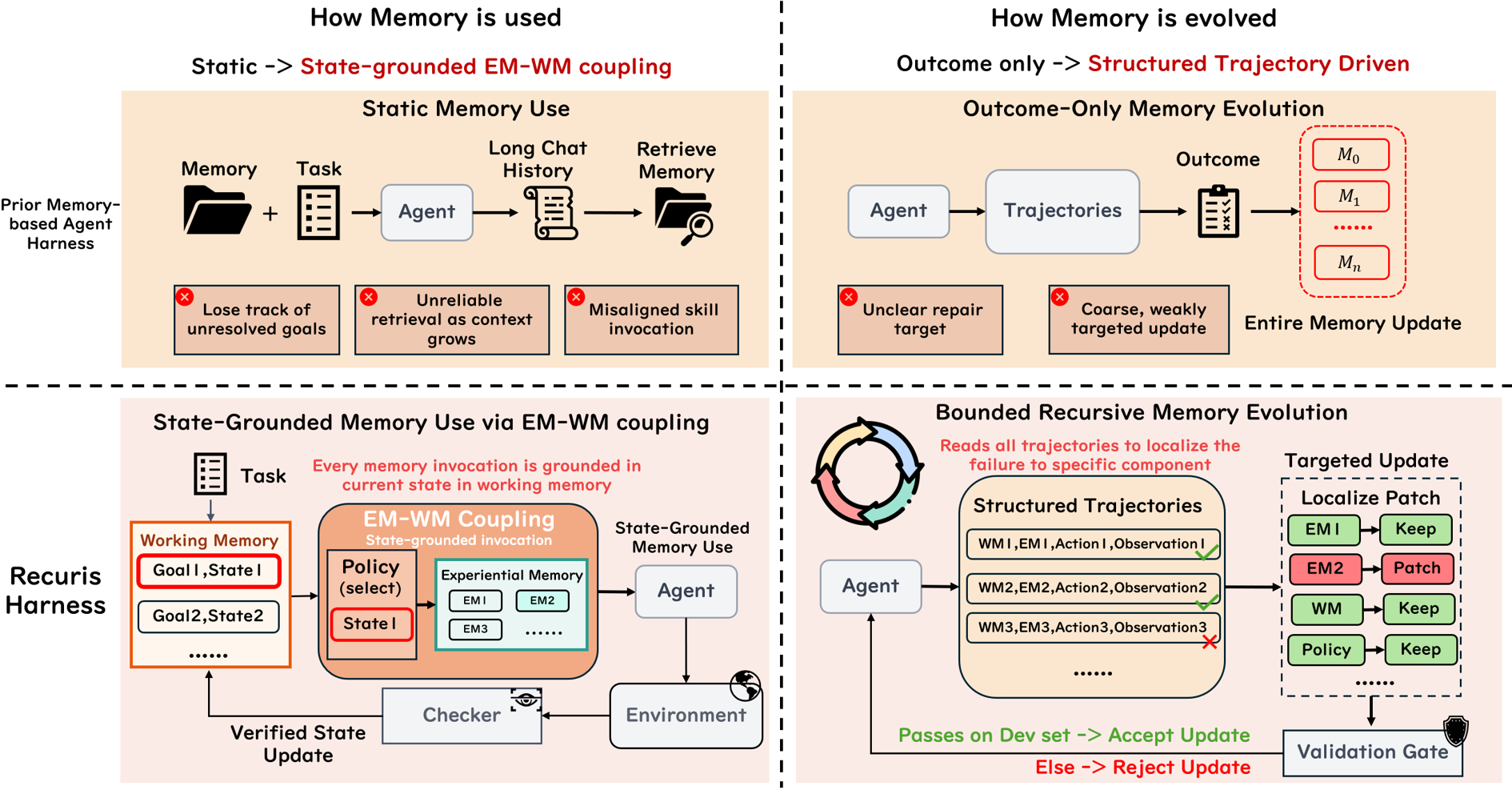}
  \caption{
    \textbf{Two shifts separate \ours{} from prior memory-based agent harnesses.}
    \textbf{How memory is used} (left): prior harnesses retrieve against a growing chat
    history and lose track of unresolved goals. \ours{} retrieves at execution events,
    and a checker turns the environment's response into a verified state update rather
    than trusting the dialogue.
    \textbf{How memory evolves} (right): prior harnesses rewrite the whole memory from
    one task outcome. \ours{} reads the structured trajectory, attributes each failure
    to a component, patches exactly the components implicated, and admits the result
    only through a validation gate on held-out tasks.
  }
  \label{fig:motivation}
\end{figure}

\section{Introduction}
\label{sec:intro}

Large language model (LLM) agents have demonstrated strong capabilities in reasoning, tool use, and autonomous execution. Increasingly, these capabilities are mediated by an \textbf{agent harness}: an external execution layer that coordinates memory, skill invocation, task-state tracking, tool interaction, and verification around the underlying model. Yet achieving \textbf{recursive self-improvement (RSI)} remains a fundamental challenge, particularly in \textbf{long-horizon tasks} where goals, observations, and failures continuously evolve throughout execution. In these settings, a capable harness must do more than provide a fixed set of tools or memories: it must maintain an accurate understanding of the current task state, invoke the appropriate experience as new requirements emerge, and turn execution failures into improvements that affect future behavior. As interaction histories grow, however, agents often lose track of unresolved goals and invoke skills that are no longer aligned with the current task state. Long-horizon execution therefore exposes a central obstacle to RSI: agents may accumulate experience, while the memory mechanisms through which the harness organizes, selects, and improves that experience remain largely fixed.

Existing experiential-memory methods attempt to address this limitation by retrieving skills from the initial instruction or the full interaction history. However, such retrieval becomes increasingly unreliable as execution unfolds: the initial instruction may no longer reflect the current problem, while the growing context mixes outdated information, completed steps, unresolved requirements, and execution noise. The central limitation is therefore not simply the absence of useful experience, but the lack of a \textbf{compact and reliable task state} that can continuously align stored experience with the agent's current execution needs.

We argue that \textbf{Working Memory (WM)} provides this missing state representation. WM continuously tracks the agent's current progress and unresolved goals, and uses them to select the most appropriate skill from \textbf{Experiential Memory (EM)}. After the skill is executed, environment feedback verifies progress and updates WM. This creates a closed loop:
\textbf{Task State}
$\rightarrow$
\textbf{Skill Selection}
$\rightarrow$
\textbf{Execution Feedback}
$\rightarrow$
\textbf{Updated Task State}.
This \textbf{Experiential--Working Memory Coupling} provides a stateful memory-control layer for the agent harness, keeping skill invocation grounded in the current task rather than the entire interaction history. WM determines what is needed now, EM provides the corresponding reusable experience, and verified feedback ensures that subsequent decisions are based on actual execution progress.

Importantly, this coupling also provides a structured substrate for \textbf{memory evolution}. Existing approaches often learn mainly from final success or failure. Such signals reveal that something went wrong, but provide little guidance about whether the most effective repair lies in the stored skill, the tracked task state, the skill-invocation mechanism, or the progress verifier. As a result, memory updates are often coarse or weakly targeted. By explicitly connecting task states, selected skills, actions, and execution outcomes, EM--WM coupling turns agent-harness execution into structured evidence about how memory affects behavior. A failed trajectory can therefore be localized to a specific memory component, enabling a scoped repair rather than rewriting the entire memory system. \Cref{fig:motivation} contrasts this with how experiential memory is used today. This
establishes the core logic of our approach:
\textbf{EM--WM Coupling}
$\rightarrow$
\textbf{Structured Evidence}
$\rightarrow$
\textbf{Failure Localization}
$\rightarrow$
\textbf{Targeted Memory Evolution}.


Based on this insight, we propose \textbf{Recuris}, a recursive Experiential--Working Memory architecture for long-horizon agent harnesses. Within each task, Recuris uses WM to maintain a verified task state and uses that state to invoke the most appropriate skill from EM. Across tasks, a fixed \textit{Meta-Agent} analyzes the resulting structured execution traces, identifies the memory component most likely to provide an effective repair, and proposes a localized update. Candidate updates are incorporated only after passing a fixed validation process.

The updated Skill Memory then changes how the harness behaves on future tasks, producing new states, skill invocations, and execution evidence that can support subsequent updates. Recuris therefore forms a \textbf{bounded, validation-gated recursive loop}: memory shapes harness behavior, harness behavior produces diagnostic evidence, and this evidence is used to evolve the memory mechanisms that shape future behavior. The underlying LLM and outer improvement procedure remain fixed; recursion occurs within the externalized memory-control layer of the agent harness.

We evaluate \ours{} on four long-horizon benchmarks covering tool-use dialogue, lifelong skill reuse, and terminal work, across ten models from a 3B open-source model to frontier models. \ours{} improves task success in 35 of the 37 completed model--benchmark pairs. On $\tau^2$-Retail it lifts \textbf{GPT-5.6 Sol} from $58.3$ to $76.1$ and \textbf{Claude Opus 5} from $72.4$ to $\mathbf{87.9}$, $9.7$ points above the best any model in our evaluation reaches on that benchmark without it; on SkillFlow it lifts Qwen3.6-27B from $42.2$ to $58.7$ and Qwen3.6-35B from $35.3$ to $48.8$. Frontier models are therefore not saturated on long-horizon tasks.

Our analyses locate where the gain comes from. The advantage grows with the interaction horizon instead of decaying, reaching $+32.2$ points on the longest tasks, and common long-horizon failures fall by up to $80\%$. Localizing a failure to the responsible memory component is far more accurate from the structured trace than from the task outcome alone, $64.8\%$ against $13.0\%$, which is what makes a scoped repair possible. Because recursion is confined to the memory-control layer, every gain is obtained with the base model left exactly as its provider shipped it.

Our contributions are summarized as follows:

\begin{itemize}

\item We establish \textbf{state-grounded memory use} as a key requirement for recursive self-improvement in long-horizon agent harnesses, where evolving task states demand continual alignment between accumulated experience and current execution needs.

\item We introduce \textbf{Recuris, a recursive Experiential--Working Memory architecture} that couples persistent experience with dynamically maintained and evidence-grounded task states, enabling adaptive skill invocation throughout long-horizon execution.

\item We show how EM--WM coupling turns harness execution into \textbf{structured diagnostic evidence}, enabling component-level failure localization and targeted evolution of the Skill Memory mechanisms that govern future behavior.

\item We evaluate \ours{} on four long-horizon benchmarks and ten models, studying execution reliability, skill invocation, failure localization, evolution stability, and cross-task transfer of evolved memory, and we observe consistent gains across model scale, including $\mathbf{+17.8}$ points for \textbf{GPT-5.6 Sol} and $\mathbf{+15.6}$ points for \textbf{Claude Opus 5} on $\tau^2$-bench.

\end{itemize}

\section{Method}
\label{sec:method}

\subsection{Problem Setup and Skill Memory Architecture}
\label{sec:problem-setup}

We consider an \textbf{agent harness} built around a frozen LLM $\pi_\theta$ and a tool set $\mathcal{T}$.
The harness mediates task-state tracking, experiential-memory access, skill invocation, tool interaction, and execution verification around the underlying model.
Given a task $x$, the agent maintains an interaction history $h_t$ and a working state $w_t$ at step $t$.
An invocation policy selects a set of experiential skills $\mathcal{E}_t$, after which the LLM produces an action and receives an environment observation:
\begin{equation}
a_t \sim \pi_\theta(\cdot \mid x,h_t,w_t,\mathcal{E}_t),
\qquad
o_t = \operatorname{Env}(a_t;\mathcal{T}).
\label{eq:agent-step}
\end{equation}
Here, $a_t$ is either a user-facing message or a tool call, and $\operatorname{Env}$ returns the next user response or the result of a tool in $\mathcal{T}$.
A run produces a raw trajectory $\tau=\{(a_t,o_t)\}_{t=1}^{L}$ and a task outcome $y\in\{0,1\}$.
\ours{} keeps $\theta$ fixed and changes future harness behavior by updating the external memory-control mechanisms that condition \cref{eq:agent-step}.

\subsubsection{Skill Memory Architecture}

At evolution round $k$, \ours{} represents the evolving Skill Memory as
\begin{equation}
\mathcal{M}_k
=
\left(
\mathcal{E}_k,
\mathcal{W}_k,
\rho_k,
\mathcal{C}_k
\right).
\label{eq:skill-memory}
\end{equation}
The experiential memory $\mathcal{E}_k$ stores reusable skills in the agent-skill format \citep{anthropic2025skills}.
The working-memory specification $\mathcal{W}_k$ defines the state schema and update proposal used to maintain its task-specific instance $w_t$.
The invocation policy $\rho_k$ decides at which execution events skills are retrieved and which entries of $\mathcal{E}_k$ then enter the context.
The checker set $\mathcal{C}_k$ tests whether observations support proposed state changes.
Together, these four components form the \textbf{evolving memory-control layer} of the agent harness and define the patch space of \ours{}.
A candidate patch revises every component that the Meta-Agent's diagnosis implicates and copies the remaining components unchanged, so scoping is enforced per edit rather than per round: a round may touch several components, but each edit is tied to a diagnosed failure and confined to the component that failure was attributed to.

The outer improvement procedure remains fixed.
The base LLM, tools, Meta-Agent, localization and patching procedures, validation gate, and harness mechanisms outside the memory-control layer do not change across rounds.
Thus, \ours{} recursively evolves $\mathcal{M}_k$ within a fixed outer agent harness rather than rewriting the underlying model or the full agent program.

\begin{figure}[t]
  \centering
  \includegraphics[width=\linewidth]{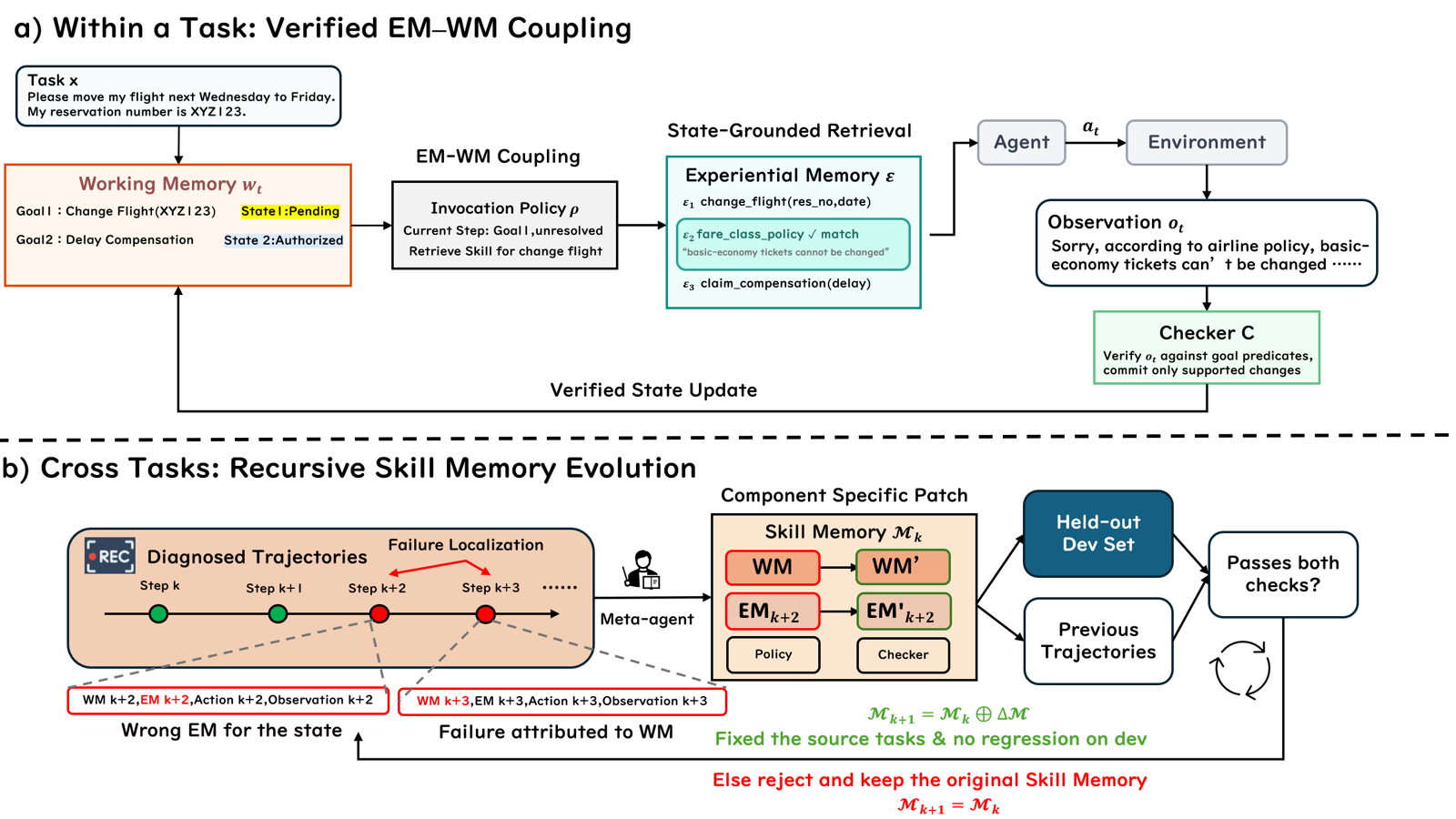}
  \caption{
    \textbf{Overview of \ours{}.}
    \textbf{(a) Within a task:} the working memory $w_t$ tracks each goal with its
    verified state. At a defined execution event the invocation policy $\rho$
    retrieves the matching skill from experiential memory $\mathcal{E}$; after the agent
    acts, the checker set $\mathcal{C}$ commits only the state changes the observation
    $o_t$ supports.
    \textbf{(b) Across tasks:} the structured trace records $(w_t, \mathcal{E}_t, a_t, o_t)$
    at every step, so a fixed Meta-Agent attributes each diagnosed failure to a
    component and patches exactly the components implicated. A fixed gate admits the
    candidate only if it repairs the source tasks without regressing a held-out
    development set; otherwise the memory is kept unchanged. The base LLM and the outer
    procedure remain fixed.
  }
  \label{fig:framework}
\end{figure}

\subsubsection{Method Overview}

As shown in \cref{fig:framework}, \ours{} has a task-level execution loop and a cross-task evolution loop.
Within a task, the current working state guides skill invocation, while execution evidence controls the next state update (\cref{sec:synergy}).
Across tasks, the fixed Meta-Agent uses failed runs to localize an effective repair target, proposes a component-specific memory patch, and relies on a fixed validation gate to decide whether the candidate should be admitted (\cref{sec:evolve}).

To connect these two loops, \ours{} records a structured execution trace at round $k$:
\begin{equation}
\Gamma_k
=
\left(
x_k,
\left\{
w_t,\mathcal{E}_t,a_t,o_t,\widetilde{w}_{t+1},c_t,w_{t+1}
\right\}_{t=1}^{L_k},
y_k
\right),
\label{eq:structured-trace}
\end{equation}
where $\widetilde{w}_{t+1}$ is the proposed state and $c_t$ contains the checker decisions made after step $t$.
For clarity, we omit the round index $k$ from variables inside a task.
Unlike the raw trajectory $\tau$, $\Gamma_k$ links each action and observation to the state that triggered skill invocation, the proposed state update, and the evidence used to accept or reject that update.
The execution loop therefore produces a structured record of how the memory-control layer influenced behavior, which the cross-task loop later uses for repair localization and memory evolution.

\subsection{Within-Task Verified EM--WM Coupling}
\label{sec:synergy}

Retrieving a skill from the initial instruction may miss needs that emerge later, while retrieving from the full interaction history must search through completed steps, outdated information, and execution noise.
\ours{} instead uses a compact working state as the interface between current progress and stored experience, allowing the harness to ground skill invocation in what remains to be accomplished.

\subsubsection{Structured Working State}

The execution harness initializes $w_0$ from the task under the specification $\mathcal{W}_k$.
Each goal entry records its content, status, supporting evidence, and an optional blocker.
The status belongs to \texttt{pending}, \texttt{done}, or \texttt{blocked}; a goal starts as \texttt{pending} unless the initial environment state already provides completion evidence.
At step $t$, $w_t$ therefore exposes what has been completed, what remains unresolved, and which observations support the recorded progress.
This compact state acts as the task-facing interface through which the harness determines what experience is relevant at the current point of execution.

\subsubsection{State-Grounded Skill Invocation}
Skill invocation decides when the agent needs external experience and which skills should enter the context.
\ours{} takes this decision at defined execution events rather than once per task:
\begin{equation}
\mathcal{E}_t
=
\rho_k(x,w_t,e_t,\mathcal{E}_k),
\qquad
\mathcal{E}_t\subseteq\mathcal{E}_k.
\label{eq:state-grounded-invocation}
\end{equation}
Here $e_t$ is the execution event that fires at step $t$, such as a drafted state-changing call or a turn boundary. Both conditioning arguments describe where the task currently stands rather than what it was asked at the outset: $w_t$ is the progress the harness has verified so far, and $e_t$ is the execution moment at which experience is needed. We call invocation \emph{state-grounded} in this sense, in contrast to retrieving once from the initial instruction or searching the full interaction history, and the deliverers below differ in which part of that execution state they read.
The invocation policy is a family of deliverers, each defined by a trigger predicate over $(w_t,e_t)$ and a retrieval key over $\mathcal{E}_k$, and a memory declares which deliverers it uses.
Two members are instantiated in this paper.

\textbf{Call-time invocation} fires when the agent drafts a state-changing tool call, and its retrieval key is the tool that call names.
The drafted call is not executed: the harness returns a synthetic not-executed result in its place, so the retrieved skill reaches the model before any state-changing action is issued, and the action that is finally executed is drafted with that skill in context.
Both $\tau^2$ domains use this deliverer.

\textbf{Boundary invocation} fires at a turn boundary under a predicate on the working state, and supplies the memory's entries without conditioning on a tool.
The Terminal-Bench 2.1 memory uses it with the predicate \texttt{first\_turn}, so the skills are supplied once at the start of an attempt.

In both cases the LLM still receives the interaction history to produce its action, but which skills reach it is decided by the harness at an execution event rather than selected by the model out of the history.
Because $\rho_k$ is itself a component of $\mathcal{M}_k$, both the trigger predicate and the retrieval key are subject to evolution.

\subsubsection{Evidence-Grounded State Update}

After the agent receives $o_t$, the working-memory specification proposes a next state.
The checker set evaluates the proposal against the returned observation, and a fixed kernel commits only the supported changes:
\begin{equation}
\begin{aligned}
\widetilde{w}_{t+1}
&= U_{\mathcal{W}_k}(w_t,a_t,o_t),\\
c_t
&= \mathcal{C}_k(w_t,\widetilde{w}_{t+1},a_t,o_t),\\
w_{t+1}
&= K(w_t,\widetilde{w}_{t+1},c_t).
\end{aligned}
\label{eq:verified-state-update}
\end{equation}
Here, $U_{\mathcal{W}_k}$ proposes state changes under $\mathcal{W}_k$, $c_t$ records the checker decisions, and $K$ is the fixed commit rule in the execution harness.
A checker $C_{k,g}\in\mathcal{C}_k$ is an explicit completion predicate for goal $g$.
It evaluates the tool or environment result rather than the model's own claim that the action succeeded.
The predicate can use a structured tool receipt or a task-specific evaluator, depending on the environment.

A goal moves from \texttt{pending} to \texttt{done} only when
$C_{k,g}(w_t,\widetilde{w}_{t+1},a_t,o_t)=1$.
Invoking a skill or attempting a tool call is not completion evidence.
If the observation does not support the proposal, the goal remains pending or becomes blocked.
If a checker rejects a valid completion, the goal remains unresolved and the decision is recorded in $c_t$; this false-pending error can later be localized to $\mathcal{C}_k$ as a candidate repair target.

The resulting task-level loop is
\begin{equation}
w_t
\;\xrightarrow{\;\rho_k\;}\;
\mathcal{E}_t
\;\xrightarrow{\;\pi_\theta,\,\mathcal{T}\;}\;
(a_t,o_t)
\;\xrightarrow{\;U_{\mathcal{W}_k},\,\mathcal{C}_k,\,K\;}\;
w_{t+1}.
\label{eq:within-task-loop}
\end{equation}
The verified state keeps unresolved needs visible, while experiential memory supplies reusable knowledge relevant to the current execution stage.
Skill-guided execution then produces the evidence required for the next state update.
Beyond improving within-task execution, this coupling creates the structured trace in \cref{eq:structured-trace}, providing the evidence that connects harness behavior to subsequent memory evolution.

\subsection{Bounded Cross-Task Skill Memory Evolution}
\label{sec:evolve}

Final task outcomes reveal whether a run succeeded or failed, but they provide little guidance about which part of the memory-control layer should be changed.
The structured trace preserves the intermediate evidence needed to identify where the working state, invoked skills, and execution results first become inconsistent.
\ours{} uses this evidence to localize a promising repair target and apply a scoped update rather than rewriting the full harness.

\subsubsection{Trace-Based Failure Localization}

For the failed runs at round $k$, the fixed localization stage of the Meta-Agent produces a diagnosis: a set of observed failures, each attributed to the component that offers the most effective repair,
\begin{equation}
D_k
=
\mathcal{A}_{\text{fixed}}(\Gamma_k,\mathcal{M}_k)
=
\left\{(f_j,z_j)\right\}_{j=1}^{J_k},
\qquad
z_j\in\left\{\mathcal{E},\mathcal{W},\rho,\mathcal{C}\right\}.
\label{eq:failure-attribution}
\end{equation}
Here $\mathcal{A}_{\text{fixed}}$ is a fixed localization procedure, $f_j$ is a diagnosed failure, and $z_j$ is the component it is attributed to; we write $Z_k=\{z_j\}_{j=1}^{J_k}$ for the set of components the diagnosis implicates, which may be a single component or several.
This is a \textbf{repair decision rather than a claim of causal identification}: multiple mechanisms may contribute to a failed trajectory, and \ours{} attributes each diagnosed failure to the component on which a localized intervention is most likely to help.
An attribution to $\mathcal{E}$ indicates a missing or flawed skill.
An attribution to $\mathcal{W}$ indicates an omitted goal, an unsuitable state field, or an incorrect state-update proposal.
An attribution to $\rho$ indicates a missed, late, or irrelevant skill invocation.
An attribution to $\mathcal{C}$ indicates that a supported transition was rejected or an unsupported transition was accepted.

\subsubsection{Component-Specific Patching}

The fixed patching stage then proposes one edit per implicated component and applies them together as a single candidate:
\begin{equation}
\Delta m_{z}
=
\mathcal{P}_{\text{fixed}}
\left(
\Gamma_k,\mathcal{M}_k,D_k,z
\right)
\;\;\text{for } z\in Z_k,
\qquad
\mathcal{M}^{+}_k
=
\mathcal{M}_k\oplus_{Z_k}\left\{\Delta m_{z}\right\}_{z\in Z_k}.
\label{eq:targeted-memory-update}
\end{equation}
The operator $\oplus_{Z_k}$ changes exactly the implicated components; every component outside $Z_k$ is copied from $\mathcal{M}_k$.
Each edit stays inside the component its failure was attributed to: a patch may add or revise a skill in $\mathcal{E}_k$, change a state field or update proposal in $\mathcal{W}_k$, adjust a trigger predicate or retrieval key in $\rho_k$, or revise a completion predicate in $\mathcal{C}_k$.
This patch space constrains self-improvement to interpretable, component-scoped changes in the memory-control layer rather than allowing a candidate to freely rewrite the full agent harness.

\subsubsection{Validation-Gated Patch Admission}

A candidate does not enter the deployed memory-control layer immediately.
The fixed gate $\mathcal{G}_{\text{fixed}}$ compares $\mathcal{M}^{+}_k$ with $\mathcal{M}_k$ on the failed task and a held-out development set.
The development set includes anchor tasks that the current memory already solves.
The gate accepts a patch only when it repairs the target failure and satisfies the preset regression criterion:
\begin{equation}
\mathcal{M}_{k+1}
=
\begin{cases}
\mathcal{M}^{+}_k,
& \text{if }\mathcal{G}_{\text{fixed}}
(\mathcal{M}^{+}_k,\mathcal{M}_k;x_k,\mathcal{D}_{\mathrm{dev}})=1,\\[3pt]
\mathcal{M}_{k},
& \text{otherwise}.
\end{cases}
\label{eq:memory-update}
\end{equation}

Here, $x_k$ is the failed source task and $\mathcal{D}_{\mathrm{dev}}$ is the held-out development set containing the anchor tasks.
The evolve split supplies trajectories for localization and patch generation.
The development split is used only by $\mathcal{G}_{\text{fixed}}$.
Test tasks, trajectories, and scores never enter localization, patch generation, or admission.
The fixed gate therefore provides an external acceptance boundary around the recursively evolving memory-control layer.

\subsubsection{Bounded Recursive Evolution}

After each admission decision, subsequent tasks run with $\mathcal{M}_{k+1}$ and produce new structured traces.
The complete loop is
\begin{equation}
\mathcal{M}_k
\rightarrow
\Gamma_k
\xrightarrow{\;\mathcal{A}_{\text{fixed}}\;}
D_k
\xrightarrow{\;\mathcal{P}_{\text{fixed}},\,\oplus_{Z_k}\;}
\mathcal{M}^{+}_k
\xrightarrow{\;\mathcal{G}_{\text{fixed}}\;}
\mathcal{M}_{k+1}
\rightarrow
\Gamma_{k+1}.
\label{eq:recursive-memory-loop}
\end{equation}

We call this process recursive because each admission decision changes the memory-control layer used for future execution, and the resulting harness behavior produces new evidence that can trigger subsequent updates.
Thus, an accepted modification does not merely repair the current failure: it changes the conditions under which later states are represented, skills are invoked, progress is verified, and future failures are observed.

The recursion is intentionally bounded.
The base LLM, Meta-Agent, localization and patching procedures, validation gate, and harness mechanisms outside $\mathcal{M}_k$ remain fixed.
\ours{} therefore realizes \textbf{recursive evolution within the externalized memory-control layer of a fixed outer agent harness}, rather than unconstrained self-modification of the underlying model or the full agent program.

\subsubsection{Test-Time Adaptation Mode}
\label{sec:tta-method}
The same machine and Meta-Agent also run in a second mode, which narrows both the evidence pool and the patch space to a single task. Given a single task and a budget of $N$ attempts, the agent executes the task and a hidden verifier returns one bit. On failure, the Meta-Agent receives the task instruction, the failed trajectory, and that bit, never the verifier, its tests, or any expected output; the restriction is enforced by the harness. The Meta-Agent proposes a localized update to the experiential memory, and the agent retries under the updated memory, stopping at the first success. Because retrying alone is a strong confound on high-variance benchmarks, we evaluate this mode against a budget-matched control that retries under the frozen initial memory, sharing its first rollout with the adaptation configuration so the contrast is paired (\cref{sec:tta}).

\section{Experiments}
\label{sec:experiments}
\subsection{Experimental Setup}
\label{sec:setup}

\hpara{Benchmarks and metrics.}
We evaluate on the two tool-use domains of $\tau^2$-Bench \citep{barres2025tau2bench}, in which an agent must satisfy a simulated user's request through policy-constrained tool calls: $\tau^2$-Retail (114 tasks) and $\tau^2$-Airline (50 tasks). An attempt counts as successful only when the environment verifier returns full reward, so an episode that ends in agreement with the user but leaves the database unchanged is a failure. Every task also carries a reference list of the actions it requires, split into reads that query the environment and writes that change it. We report \textit{read-action recall} and \textit{required-write recall} separately, because together they separate knowing what to do from carrying it out. We also evaluate on \textbf{SkillFlow}~\citep{skillflow}, a benchmark for lifelong skill discovery in which 166 tasks are organised into 20 families, and in which the tasks inside a family are constructed to share one execution flow. It is the one benchmark here whose structure is procedural by design, which is what makes it the sharpest test of whether an evolved package carries procedure. Every task ships its own verifier script and returns a binary reward, so SkillFlow is scored programmatically rather than by a model judge. Terminal-Bench 2.1 is introduced with the adaptation experiment it supports (\cref{sec:tta}).

\hpara{Models.}
Every configuration runs a single frozen instruction-tuned model as the agent at temperature $0$; on $\tau^2$-Bench, whose environment is dual-control, that same model also plays the simulated user. Unless a row names another, that model is \texttt{doubao-\allowbreak seed-\allowbreak 2-0-\allowbreak pro}, which we call the \textit{deployment model}: it is the model the evolution loop runs on, and the model every configuration in a comparison executes on. It is deliberately mid-sized rather than frontier. The target models in \cref{tab:main} span three open-weight families, Qwen3~\citep{qwen3tech}, Granite~\citep{granite4} and gpt-oss~\citep{gptoss}, and three frontier models reached through a single proxy. No model weights are updated anywhere in this paper. Each task is attempted $k$ times independently, $k=4$ unless stated otherwise, and each attempt is capped at 200 steps and 10 consecutive tool errors.

\hpara{How the Skill Memory is built.}
For each benchmark we build one Skill Memory, and we build it on the deployment model alone. That model runs the benchmark's own reference agent, unmodified, the $\tau^2$-Bench tool-calling agent, the Qwen-Code CLI agent on SkillFlow and Terminus-2 on Terminal-Bench 2.1, and the failures it produces are the only evidence the loop ever sees. A fixed \textit{Meta-Agent}, itself an LLM agent rather than a hand-written procedure, reads the structured traces of those failed episodes, attributes each diagnosed failure to a memory component, and writes the component-scoped patch that the gate admits or rejects. It is built on Claude Code, is never modified, and never sees the test split: the same implementation, prompts and procedure run at every round and on every benchmark, which is what \textit{fixed} means throughout this paper. The memory a benchmark ends up with is therefore shaped by what a mid-sized model gets wrong, not by what a stronger one would. Nothing in the loop is specific to that model, and running it on another would yield a memory specific to that one; we report the single-source setting deliberately, because a memory tuned per model could absorb that model's idiosyncrasies whereas a single memory must carry what the models share, and because evolution is the expensive step and a memory evolved once costs nothing to reuse. The splits that separate evolution, gating and held-out evaluation are fixed before any run and given in \cref{app:splits}.

\hpara{How the Skill Memory is evaluated.}
That memory is then loaded into \ours{} and used directly at inference: nothing is evolved during evaluation, and the base model of every row stays frozen. We compare three configurations. The \textit{base agent} is the benchmark's reference implementation, run as above. \ours{} \textit{with initial memory} carries $\mathcal{M}_0$, the neutral memory the loop begins from, and \ours{} \textit{with evolved memory} carries what the loop produced; both add the memory-control layer inside that same reference agent and change nothing else. Within a comparison the configurations share the model, the tool set, the task set, the seed assignment and the budget, and on $\tau^2$-Bench the user simulator as well, so a gain cannot come from a harness chosen in our own favour. What the layer contributes before anything is learned is measured rather than assumed, and is indistinguishable from zero on four target models (\cref{app:harness-ablation}), so what sits above $\mathcal{M}_0$ belongs to the evolution. Every model in \cref{tab:main} other than the deployment model receives that same memory unchanged; \cref{sec:transfer-models} reports what it is worth to them.

\subsection{Overall Performance}
\label{sec:overall}
\begin{table}[t]
\centering
\caption{%
  \textbf{\ours{} improves task success where the task family has shared structure,
  and its value does not follow model scale.}
  Cells are \texttt{avg@4} task success (\%); higher is better. Each model contributes
  a pair of rows, the benchmark's own reference agent alone and that same agent with
  \ours{} (grey). \textbf{Bold} marks the better of each pair; the subscript is $\Delta$,
  shaded by the size of the gain; $^{\dagger}$ marks a paired task-clustered bootstrap
  95\% CI excluding zero. Agents, models and protocol are given in \cref{sec:setup}.
  \na{}~not evaluated on that benchmark.
}
\label{tab:main}
\small
\setlength{\tabcolsep}{4pt}
\renewcommand{\arraystretch}{1.05}
\begin{tabular}{@{}l cccc@{}}
\toprule
& \multicolumn{3}{c}{\textit{Cross-task evolution}} & \textit{Within-task adaptation} \\
\cmidrule(lr){2-4}\cmidrule(lr){5-5}
\textbf{Model} & \textbf{$\tau^2$-Retail} & \textbf{$\tau^2$-Airline} & \textbf{SkillFlow} & \textbf{Terminal-Bench 2.1} \\
\midrule
\multicolumn{5}{@{}l}{\itshape Open-weight models} \\
Granite-4.1-3B & 9.7 & 34.3 & \best{0.3} & 0.6 \\
\oursrow
\quad $+$ \ours{} & \best{23.0}\,\gaindag{13.4} & \best{39.8}\,\gain{5.5} & 0.0\,\drop{0.3} & \best{3.1}\,\phgain{2.5} \\
\addlinespace[1.5pt]
Qwen3.5-4B & 68.0 & 75.3 & 6.0 & 10.1 \\
\oursrow
\quad $+$ \ours{} & \best{68.3}\,\gain{0.3} & \best{79.0}\,\gain{3.8} & \best{7.1}\,\gain{1.1} & \best{13.0}\,\phgain{2.9} \\
\addlinespace[1.5pt]
Qwen3.5-9B & 77.6 & 75.5 & 15.1 & 17.4 \\
\oursrow
\quad $+$ \ours{} & \best{79.6}\,\gain{2.0} & \best{78.4}\,\gain{2.9} & \best{18.4}\,\gain{3.4} & \best{20.5}\,\phgain{3.1} \\
\addlinespace[1.5pt]
GPT-OSS-20B & 50.6 & 54.8 & 7.8 & 3.9 \\
\oursrow
\quad $+$ \ours{} & \best{60.8}\,\gaindag{10.2} & \best{59.3}\,\gaindag{4.5} & \best{10.4}\,\gaindag{2.6} & \best{6.7}\,\phgain{2.8} \\
\addlinespace[1.5pt]
Qwen3.6-27B & 62.8 & 79.0 & 42.2 & 38.8 \\
\oursrow
\quad $+$ \ours{} & \best{71.2}\,\gaindag{8.3} & \best{80.0}\,\gain{1.0} & \best{58.7}\,\gaindag{16.6} & \best{42.1}\,\phgain{3.3} \\
\addlinespace[1.5pt]
Qwen3.6-35B & 78.2 & 80.3 & 35.3 & 33.1 \\
\oursrow
\quad $+$ \ours{} & \best{78.5}\,\gain{0.3} & \best{81.5}\,\gain{1.3} & \best{48.8}\,\gaindag{13.5} & \best{36.4}\,\phgain{3.3} \\
\midrule
\multicolumn{5}{@{}l}{\itshape Frontier models} \\
Gemini 3.7 Flash & 73.5 & \best{86.5} & \na & 79.8 \\
\oursrow
\quad $+$ \ours{} & \best{78.3}\,\gain{4.8} & 85.0\,\drop{1.5} & \na & \best{82.4}\,\phgain{2.6} \\
\addlinespace[1.5pt]
GPT-5.6 Sol & 58.3 & 79.0 & \na & 83.2 \\
\oursrow
\quad $+$ \ours{} & \best{76.1}\,\gaindag{17.8} & \best{86.0}\,\gaindag{7.0} & \na & \best{86.4}\,\phgain{3.2} \\
\addlinespace[1.5pt]
Claude Opus 5 & 72.4 & 89.5 & \na & 84.6 \\
\oursrow
\quad $+$ \ours{} & \best{87.9}\,\gaindag{15.6} & \best{90.5}\,\gain{1.0} & \na & \best{88.4}\,\phgain{3.8} \\
\addlinespace[1.5pt]
Doubao-2.0-Pro (deployment) & 58.1 & 75.5 & 34.6 & 46.1 \\
\oursrow
\quad $+$ \ours{} & \best{81.4}\,\gaindag{23.3} & \best{80.5}\,\gain{5.0} & \best{51.4}\,\gaindag{16.8} & \best{48.9}\,\gain{2.9} \\
\bottomrule
\end{tabular}
\end{table}

For each benchmark, the Skill Memory evaluated here is the one the evolution loop of \cref{sec:evolve} produced on the deployment model, \texttt{doubao-\allowbreak seed-\allowbreak 2-0-\allowbreak pro}. That memory is then loaded into \ours{} and used directly at inference: nothing is evolved during evaluation, and the base model of every row stays frozen.

\Cref{tab:main} presents the main results: four benchmarks under two adaptation regimes, with each model evaluated with and without \ours{}. \ours{} improves task success in 35 of the 37 completed model--benchmark pairs. The largest gains appear on $\tau^2$-Retail and SkillFlow, where \ours{} improves the deployment model by 23.3 and 16.8 points, reaching $81.4\%$ and $51.4\%$, both with intervals excluding zero. The gain belongs to memory that revises itself rather than to memory as such: a Skill Memory held fixed adds nothing an interval separates from zero, which we take up in \cref{sec:ablation-coupling}. The gains are also not simply more context or more compute: the base model is frozen, every comparison shares its interaction budget, and on Terminal-Bench 2.1 both configurations run the same four rollouts. Where we can vary context directly the extra context hurts: a regime that keeps the whole skill library standing in the prompt carries $3{,}111$ more tokens at the first call than \ours{}, scores 18 points lower, and costs $46\%$ more per success (\cref{app:compute}).

\hpara{Task structure decides the adaptation regime.}
What decides it is shared structure, not difficulty. $\tau^2$-Retail, $\tau^2$-Airline and SkillFlow share tools, policies and, in SkillFlow, whole task families, so a repair earned on one split is worth carrying to another and memory can be evolved across tasks. Terminal-Bench 2.1 has no such structure, and cross-task evolution on it admitted no patch in thirteen runs of the evolution loop. What works there is retrying inside a single task, with or without memory written between attempts; \cref{sec:tta} decomposes the two and reports where adaptation adds its margin.

\subsection{Analysis of EM--WM Coupling in Long Horizon Tasks}

A long-horizon deficit can be a failure to find the right knowledge or a failure to act on it. We first separate the two, then ask which memory prevents the one that dominates, and finally which mechanism inside the harness supplies it. That last answer, which turns out to differ by domain, is what motivates \cref{sec:evolve}.

\subsubsection{\ours{} Remains Reliable over Longer Interaction Horizons}
\label{sec:long-horizon}
%

\begin{figure}[t]
  \centering
  \includegraphics[width=\linewidth]{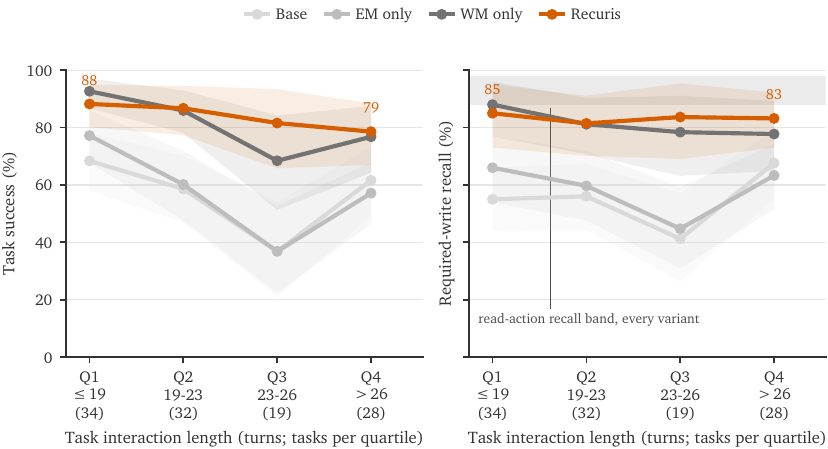}
  \caption{
    \textbf{Long-horizon failure is an execution problem, not a retrieval problem.}
    Task success (left) and required-write recall (right) for the four ablation variants on
    $\tau^2$-Retail, stratified into quartiles by the number of turns a task takes to
    complete; quartile sizes appear under each tick and the grey band spans read-action
    recall for every variant in every quartile.
    Shaded regions are paired task-clustered bootstrap $95\%$ CIs ($10{,}000$ resamples of
    tasks).
    Quartiles are defined by task-intrinsic completion length, never by the length a variant
    realises, which is an outcome of the variant.
  }
  \label{fig:long-horizon}
\end{figure}

\ours{}'s advantage does not shrink as tasks get longer. We stratify $\tau^2$-Retail into quartiles by how long a task intrinsically takes to finish, taken as the per-task median horizon over the passing episodes of all variants, so that one task-to-quartile map holds for every variant.\footnote{Stratifying on the horizon a variant actually realises would be circular: an agent that quits early produces a short episode, and that choice alone moves 44 of the 114 tasks between quartiles.} \ours{} leads the base agent in all four quartiles, by between $+17.0$ and $+44.7$ points, with no monotone decline in length (\cref{fig:long-horizon}).

The reason is that length does not break retrieval. Read-action recall, the share of the queries a task requires that the agent actually issues, stays within $88.0$--$97.9\%$ for every variant in every quartile: no variant, at any length, fails to find out what its task needs. The separation is entirely on the write path, where \ours{} exceeds the base agent by $26.7$ points of required-write recall.

Length therefore costs completion, not comprehension, and the shortfall is concrete. The base agent ends $42\%$ of the episodes that require a write having executed none of them, against $16\%$ for \ours{}, while the median turn of the first correct write is identical across all four variants: what changes is not when the agent acts but whether it acts at all. This is a stratified re-analysis of held-out episodes rather than a controlled manipulation of horizon, so we read it as a consistent pattern. It is also why a memory that tracks what remains outstanding should beat one that only supplies what to do, which is the comparison \cref{sec:ablation-coupling} makes directly.
\FloatBarrier
\subsubsection{EM--WM Coupling Improves Long-Horizon Execution}
\label{sec:ablation-coupling}

\begin{table}[tb]
\centering
\caption{
  \textbf{Coupling both memories gives the strongest variant in each domain.}
  Each variant removes one component from one fixed Skill Memory; variants within a domain
  are matched on model, user simulator, skill library, tools, tasks, seeds and budget.
  Cells give success rate (successful/total episodes); $\Delta$ is against that domain's
  base agent (\cref{app:stats}) and $^{\dagger}$ marks an interval excluding zero.
  The two domains were evaluated separately, so values compare within a column; the
  model-controlled airline cells are omitted because that batch cannot be paired.
}
\label{tab:ablation-coupling}
\small
\setlength{\tabcolsep}{5pt}
\begin{tabular}{lcccc}
\toprule
& \multicolumn{2}{c}{$\tau^2$-Retail (114 tasks, 456 episodes)} & \multicolumn{2}{c}{$\tau^2$-Airline (50 tasks, 200 episodes)} \\
\cmidrule(lr){2-3}\cmidrule(lr){4-5}
Variant & Success ($\uparrow$) & $\Delta$ \tiny{[95\% CI]} & Success ($\uparrow$) & $\Delta$ \tiny{[95\% CI]} \\
\midrule
Base (no EM, no WM) & 58.1 \tiny{(265/456)} & \na & 75.5 \tiny{(151/200)} & \na \\
EM only & 60.1 \tiny{(274/456)} & $+2.0$ \tiny{[$-4.0$,\,$+7.9$]} & 77.0 \tiny{(154/200)} & $+1.5$ \tiny{[$-4.5$,\,$+7.5$]} \\
WM only & \sbest{82.0} \tiny{(374/456)} & $+23.9^{\dagger}$ \tiny{[$+17.5$,\,$+30.3$]} & \sbest{79.5} \tiny{(159/200)} & $+4.0$ \tiny{[$-5.0$,\,$+13.5$]} \\
Model-controlled invocation & 65.6 \tiny{(299/456)} & $+7.5^{\dagger}$ \tiny{[$+1.5$,\,$+13.4$]} & \na & \na \\
\oursrow
\textbf{EM\,+\,WM (\ours{})} & \best{83.6} \tiny{(381/456)} & $\mathbf{+25.4}^{\dagger}$ \tiny{[$+18.4$,\,$+32.5$]} & \best{84.0} \tiny{(168/200)} & $\mathbf{+8.5}$ \tiny{[$-1.0$,\,$+18.5$]} \\
\bottomrule
\end{tabular}
\end{table}

\begin{figure}[t]
  \centering
  \includegraphics[width=\linewidth]{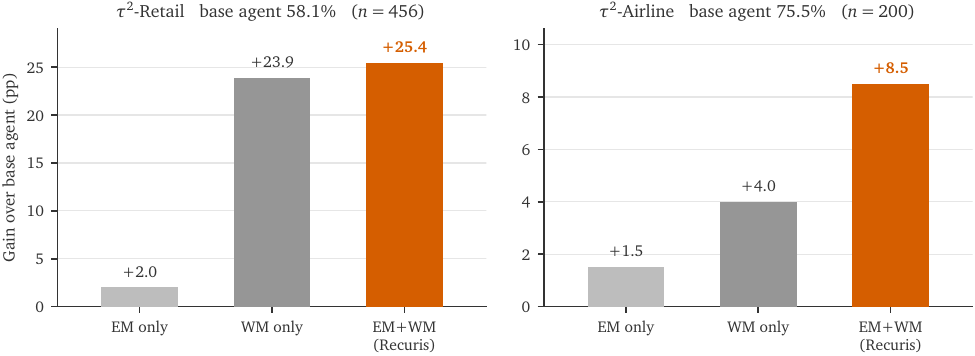}
  \caption{
    \textbf{Coupling of EM and WM.}
    Bars give each variant's gain over the base agent of the same domain, whose absolute
    success rate is stated above each panel, so the base agent is the zero of this chart
    and the axis is not truncated.
    Confidence intervals are reported in \cref{tab:ablation-coupling} rather than drawn
    here; on $\tau^2$-Airline every interval includes zero, so that panel shows a
    direction and not an established effect.
    Panels use separate scales because the two domains were evaluated in separate runs.
  }
  \label{fig:ablation-coupling}
\end{figure}

Which memory prevents the unexecuted write? We compare five configurations that differ only in which memory is present and how skills reach the model: a base agent with neither memory, an EM-only variant that injects skills with no working state to select against, a WM-only variant that tracks and verifies the state with no experiential memory, the full coupled system, and a model-controlled variant that holds the same skill library as \ours{} but injects all of it every turn and leaves the model to decide when to use it.

The working state carries the level. Against the base agent, adding experiential memory is worth $+2.0$ points on $\tau^2$-Retail and $+1.5$ on $\tau^2$-Airline, neither excluding zero, while adding a verified working state is worth $+23.9$. Against a working state alone, experiential memory adds a further $+1.5$ with an interval of $[-2.4, +5.7]$, so the skills do not carry capability on their own either.

What the skills do carry becomes visible only against invocation. The model-controlled variant holds the same library and puts strictly more skill text in front of the model than \ours{} does, and \ours{} scores $18.0$ points above it, with an interval of $[+11.6, +24.6]$ that excludes zero. Availability is not the mechanism; the state that decides when to invoke is. Experiential memory is \emph{invocation-conditional}: what it is worth depends less on what it contains than on whether anything knows when it is needed.

$\tau^2$-Airline resolves less than $\tau^2$-Retail: with 50 tasks every interval in its column includes zero, so we read it as a direction rather than an effect. In that direction the two domains ask different things of the state. $\tau^2$-Airline tasks hinge on policy constraints that a skill states but that the agent needs only once a goal becomes blocked, so the state has to be correct for the right skill to be reachable at all. $\tau^2$-Retail tasks more often fail by leaving a tracked write unexecuted, which a verified state catches without consulting a skill.
\FloatBarrier
\subsubsection{The Critical Memory Component Differs across Domains}
\label{sec:mechanism-ablation}

Which mechanism is critical is a property of the domain, not of the architecture. The verified working state is maintained by four mechanisms acting at different points in a turn: write review inspects a state-changing action before it executes, the truth guard audits a completion claim after the fact, the status board governs how the state is presented, and gate termination governs when an episode may end. We remove each in turn and re-run both domains.

\begin{figure}[htbp]
  \centering
  \includegraphics[width=\linewidth]{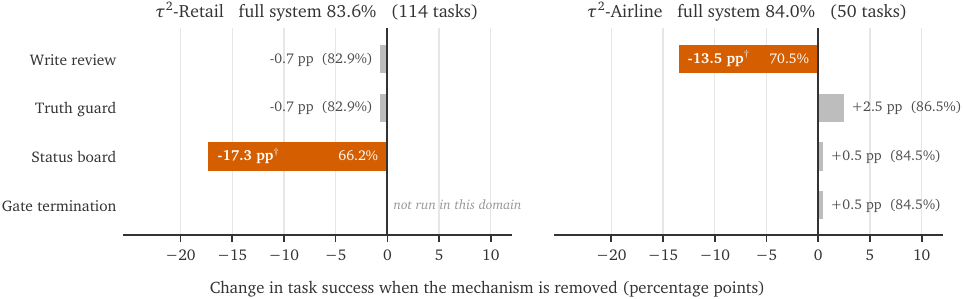}
  \caption{
    \textbf{The critical mechanism differs between domains.}
    Each bar removes one harness mechanism from the full system and reports the change in
    task success against that domain's full system. The two mechanisms that matter are
    disjoint: write review carries $\tau^2$-Airline and is inert on $\tau^2$-Retail, while
    the status board does the reverse.
    $^{\dagger}$ marks a paired task-clustered bootstrap 95\% CI that excludes zero:
    $[-22.6, -12.3]$ for the status board on $\tau^2$-Retail and $[-21.0, -6.5]$ for write
    review on $\tau^2$-Airline. Every other interval includes zero.
    Gate termination was not ablated on $\tau^2$-Retail.
  }
  \label{fig:mechanism-ablation}
\end{figure}

Exactly one mechanism is critical in each domain, and it is not the same one (\cref{fig:mechanism-ablation}). On $\tau^2$-Airline, removing write review costs $13.5$ points while the status board costs nothing measurable; on $\tau^2$-Retail the pattern reverses, the status board costing $17.3$ points and write review $0.7$ with an interval including zero. This is a double dissociation, not a difference in how well the two domains resolve effects.

What separates the mechanisms that matter from the one that does not is \emph{when} they act. The truth guard is the only one that checks after the fact, and the only one that never matters, despite rejecting 172 unsupported completion claims: a write that has executed has already moved the environment, and a later audit can record the error but not undo it.

If the mechanism that carries a domain cannot be read off the architecture, it cannot be chosen at design time either, and any fixed allocation of effort across components will be wrong somewhere. That is what \cref{sec:evolve} is for: rather than committing in advance, \ours{} reads the repair target off the trace of the run that failed. 

\FloatBarrier
\subsubsection{State-Grounded Skill Invocation Is More Accurate and Efficient}
\label{sec:invocation}

The value of a skill library depends on when its entries reach the model. We hold the library fixed and vary who decides those moments, across four configurations on $\tau^2$-Retail matched on model, decoding, harness version and seeds: the base agent; a working-memory configuration with the experiential skills removed; a model-controlled variant that injects all eight tool skills into context on every turn and leaves the model to decide when to use them; and \ours{}, which supplies the one skill matched to each drafted state-changing call. The last two carry the same ten-skill library with byte-identical bodies, so what separates them is invocation control and nothing else.

\begin{table}[t]
\centering
\caption{%
  \textbf{Control over skill invocation matters more than skill content.}
  $\tau^2$-Retail, 114 tasks, $k=4$, 456 episodes per variant at matched model, decoding and
  harness. Model-controlled and \ours{} carry the same ten skills, byte-identical.
  Required-write recall is the share of the reference plan's database-mutating actions the
  agent performed; \emph{omitted} counts those never issued at all. Cost charges every
  episode, including failures, to the variant that spent it.
}
\label{tab:invocation}
\small
\setlength{\tabcolsep}{8pt}
\begin{tabular}{@{}l cccc@{}}
\toprule
& \textbf{Base} & \textbf{WM only} & \textbf{Model-controlled} & \textbf{\ours{}} \\
\midrule
Task success (\%) & 58.1 & 82.0 & 65.6 & \best{83.6} \\
Required-write recall (\%) & 55.7 & 80.9 & 61.1 & \best{82.4} \\
Omitted required writes / episode & 0.596 & 0.145 & 0.417 & \best{0.121} \\
Agent tokens per success (k) & 116 & 102 & 147 & \best{101} \\
\bottomrule
\end{tabular}
\end{table}

\begin{figure}[t]
  \centering
  \includegraphics[width=\linewidth]{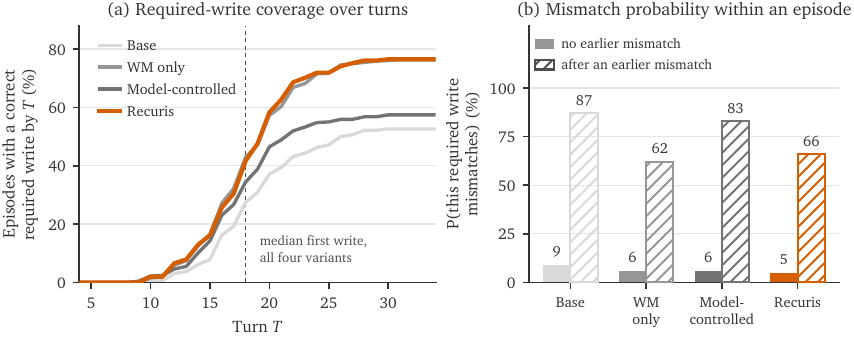}
  \caption{%
    \textbf{State-grounded invocation recovers required writes the agent would otherwise omit, and
    enters the error-compounding regime least often.}
    (a) share of all 456 episodes that have issued a correct required write by turn $T$; the
    dashed rule marks the median turn of the first correct write, which is identical in all
    four variants.
    (b) probability that a required write mismatches, split by whether an earlier write in the
    same episode already mismatched.
  }
  \label{fig:invocation}
\end{figure}

Invocation control, not skill content, drives the result, and it drives both halves of it. \ours{} performs $82.4\%$ of the required database writes against $61.1\%$ under model control, and the same ordering carries through to task success, $83.6$ against $65.6$ (\cref{tab:invocation}). Putting the whole library in context and delegating invocation to the model scores below the identical configuration carrying no skills at all, $82.0$, and costs more to run: $147$k agent tokens per success against $101$k for \ours{}. More skill text in front of the model is worse than none, because what the model-controlled regime lacks is a signal for \emph{when} a skill is relevant.

The advantage takes the form of coverage rather than speed. The median turn of the first correct write is 18 in all four configurations, so a skill invoked at write time does not make the agent decide sooner; what it changes is whether the write is ever issued (\cref{fig:invocation}a). Errors also compound within an episode: once a required write has mismatched, the probability that a later one mismatches exceeds $60\%$ in every configuration, and \ours{} reaches that condition least often (\cref{fig:invocation}b). Invoking the right skill at the moment of the write therefore prevents one early error from corrupting the rest of the episode.
\subsubsection{Case Study: Verified EM--WM Coupling}
\begin{figure}[!tbp][H]
  \centering
  \includegraphics[width=\linewidth]{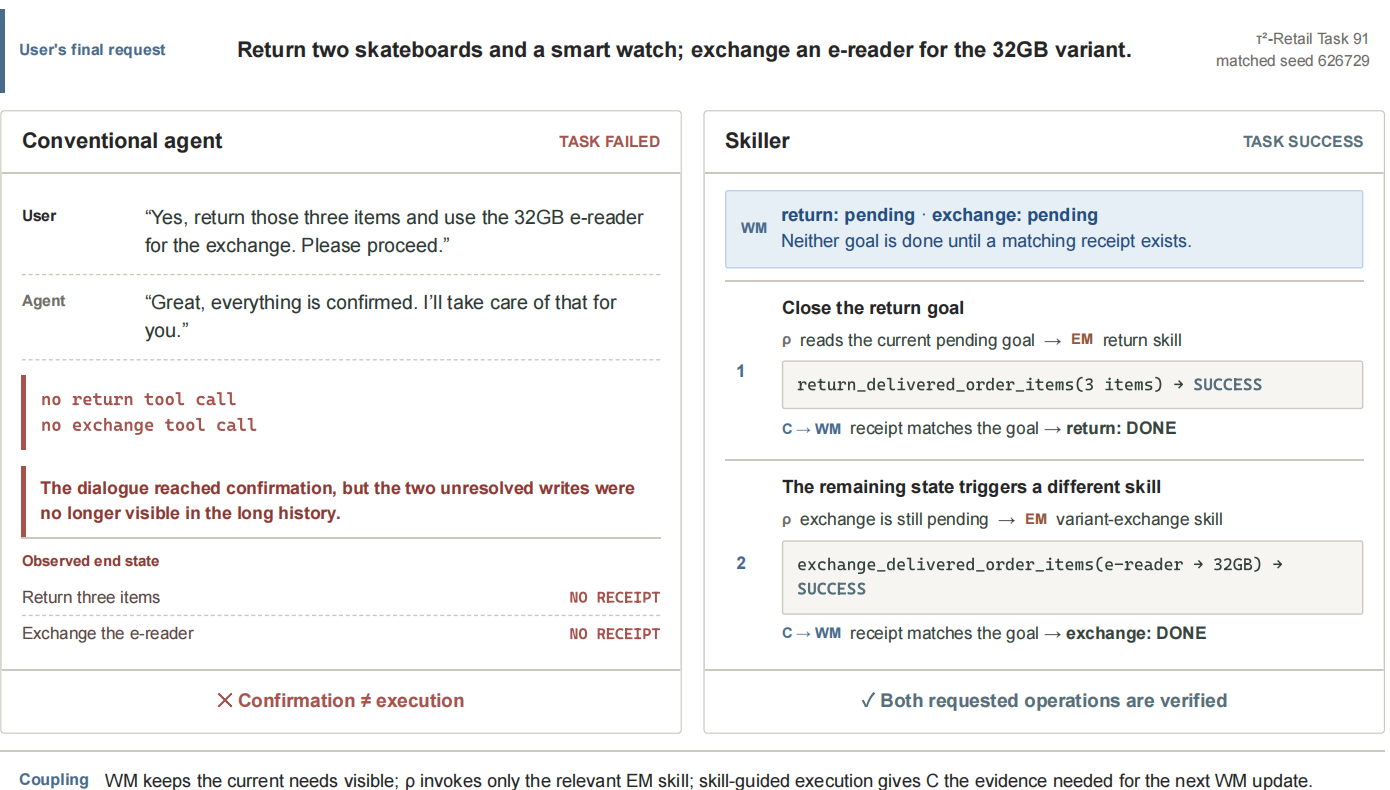}
  \caption{
    \textbf{Case study of verified EM--WM coupling on $\tau^2$-Retail Task~91.}
  }
  \label{fig:case-inference}
\end{figure}

As shown in \cref{fig:case-inference}, verbal confirmation does not close either goal. \ours{} retains both goals in working memory until skill-guided tool calls return matching receipts.  A conventional agent reaches verbal confirmation but issues neither required tool call, leaving both database updates unresolved (left). \ours{} keeps the return and exchange goals pending in working memory; $\rho$ invokes the corresponding experiential skill for each unresolved goal, and $\mathcal{C}$ marks a goal complete only after a matching successful tool receipt (right). This matched example illustrates the mechanism and is not an aggregate performance claim.

\subsection{Analysis of Recursive Skill Memory Evolution}

\subsubsection{Structured Traces Improve Failure Localization}
\label{sec:attribution}

The loop can only repair what it can localize, so we intervene on the memory rather than only observing it. We inject a known fault into one component of a working package and ask a fixed judge which component is at fault from one of three evidence conditions: the outcome alone, the raw trajectory, or the structured trace $\Gamma$ the harness emits. The pool is balanced by construction, so a judge that always names the same component scores $33.3\%$. Macro accuracy is $13.0\%$ from the outcome, $37.0\%$ from the raw trajectory and $64.8\%$ from $\Gamma$ (\cref{tab:attribution}): the outcome condition sits below the constant-answer floor, the raw trajectory clears it by less than four points, and $\Gamma$ nearly doubles it.

\begin{table}[t]
\centering
\caption{%
  \textbf{Structured traces make component faults observable, and the gain is concentrated on the
  faults the transcript cannot show.}
  A known fault is injected into one component of a working package, a skill ($\mathcal{E}$), the working-memory specification ($\mathcal{W}$), the invocation policy ($\rho$), or the checkers ($\mathcal{C}$) of \cref{eq:skill-memory}; a fixed judge names the
  component at fault from one of three evidence conditions. Nine admitted cases per class, two
  repeats, 54 verdicts per condition. Cells are per-class recall (\%); \emph{Macro} averages the
  three classes against a $33.3\%$ constant-answer floor.
}
\label{tab:attribution}
\small
\setlength{\tabcolsep}{7pt}
\begin{tabular}{@{}l ccc cc@{}}
\toprule
& \multicolumn{3}{c}{\textbf{Injected fault}} & & \\
\cmidrule(lr){2-4}
\textbf{Evidence given to the judge} & $\mathcal{E}$ & $\mathcal{W}$ & $\rho$ & \textbf{Macro} & \textbf{Macro-F1} \\
\midrule
Outcome only & 0.0 & 38.9 & 0.0 & 13.0 & 10.4 \\
Raw trajectory & 61.1 & 50.0 & 0.0 & 37.0 & 31.2 \\
Structured trace $\Gamma$ & \best{72.2} & \best{83.3} & \best{38.9} & \best{64.8} & \best{63.4} \\
\bottomrule
\end{tabular}
\end{table}

The per-fault rows say where the gain comes from, and it is observability rather than reasoning. An invocation fault is a non-event: nothing appears in the transcript where a skill should have been injected, and neither condition without $\Gamma$'s mechanism events names it even once, against $38.9\%$ from $\Gamma$. A corrupted working-memory record is visible in $\Gamma$'s state timeline and not in the dialogue, and reading that timeline lifts the class from $50.0\%$ to $83.3\%$. Skill-content faults gain least, $61.1\%$ to $72.2\%$, which is what the design predicts, since a wrong argument already sits in the transcript. Precision moves further than recall: macro precision rises from $27.6\%$ to $64.4\%$, because $\Gamma$ stops the judge charging the skills for another component's fault.

That is exactly what the loop needs, since a repair aimed at the wrong component is wasted. One of the four components is left out of the ground truth: removing the checkers changes the pass rate by zero on $\tau^2$-Retail, so their admitted cases are decode noise rather than caused failures, and testing that class needs a domain where the check binds.

\subsubsection{Recursive Evolution Yields Consistent Held-Out Gains}
\label{sec:evolution-dynamics}
%

The evolution loop converts a handful of failed training tasks into gains that hold up on tasks it never saw, and it does so consistently: every evolution run's evolved memory clears $\mathcal{M}_0$ with an interval excluding zero, across two Meta-Agent implementations (\cref{sec:metaagent-swap}) and three admission-threshold settings. An evolution run is one pass of the loop of \cref{sec:evolve}, and every package below is re-evaluated after all runs ended, on the same frozen 86-task split the Meta-Agent never reads, so the numbers sit on one scale. Two runs of $\mathcal{M}_0$ itself differ by $+0.00$ with an interval of $[-6.98, +7.27]$, which sets the resolution against which the rest is read.

\begin{table}[t]
\centering
\caption{%
  \textbf{Recursive evolution yields consistent held-out gains, and a second round compounds them.}
  Every package is evaluated on the same 86 tasks held out from the Meta-Agent at $k=4$.
  Run A's Meta-Agent runs on Claude Code, Runs B and C on DeepSeek Harness under the
  identical protocol; only Run C admits candidates inside the loop. $\Delta$ is against
  $\mathcal{M}_0$ (\cref{app:stats}); $^{\dagger}$ marks an interval excluding zero.
  \emph{Reach} is the share of held-out tasks on which the memory was invoked at least
  once, and is blank where no instrumented run exists.
}
\label{tab:evolution-dynamics}
\small
\setlength{\tabcolsep}{6pt}
\begin{tabular}{@{}ll ccc@{}}
\toprule
\textbf{Evolution run} & \textbf{Package} & \textbf{Success} & \textbf{$\Delta$ vs.\ $\mathcal{M}_0$} & \textbf{Reach} \\
\midrule
\multicolumn{2}{@{}l}{$\mathcal{M}_0$ (shared starting memory)} & 54.07 & \na & 0/86 \\
\midrule
\multirow{3}{*}{Run A} & round 1 & 63.08 & \hc{9.01}{$+9.01^{\dagger}$} \tiny{[$+1.45$,\,$+16.57$]} & 83/86 \\
 & round 2 & 65.99 & \hc{11.92}{$+11.92^{\dagger}$} \tiny{[$+4.65$,\,$+19.19$]} & 78/86 \\
 & round 3 & 65.99 & \hc{11.92}{$+11.92^{\dagger}$} \tiny{[$+4.65$,\,$+18.90$]} & -- \\
\midrule
\multirow{3}{*}{Run B} & round 1 & 64.53 & \hc{10.47}{$+10.47^{\dagger}$} \tiny{[$+3.78$,\,$+17.15$]} & -- \\
 & round 2 & 64.83 & \hc{10.76}{$+10.76^{\dagger}$} \tiny{[$+3.20$,\,$+18.31$]} & 86/86 \\
 & round 4 & 57.85 & \hc{3.78}{$+3.78$} \tiny{[$-2.33$,\,$+9.88$]} & 0/86 \\
\midrule
\multirow{3}{*}{\makecell[l]{Run C\\\itshape admitted}} & $\mathcal{M}_1$ & 64.53 & \hc{10.47}{$+10.47^{\dagger}$} \tiny{[$+3.49$,\,$+17.44$]} & 84/86 \\
 & $\mathcal{M}_2$ & \best{71.51} & \hc{17.44}{$\mathbf{+17.44}^{\dagger}$} \tiny{[$+10.47$,\,$+24.42$]} & 84/86 \\
 & final & 63.37 & \hc{9.30}{$+9.30^{\dagger}$} \tiny{[$+1.45$,\,$+17.44$]} & -- \\
\bottomrule
\end{tabular}
\end{table}

\begin{figure}[t]
  \centering
  \includegraphics[width=\linewidth]{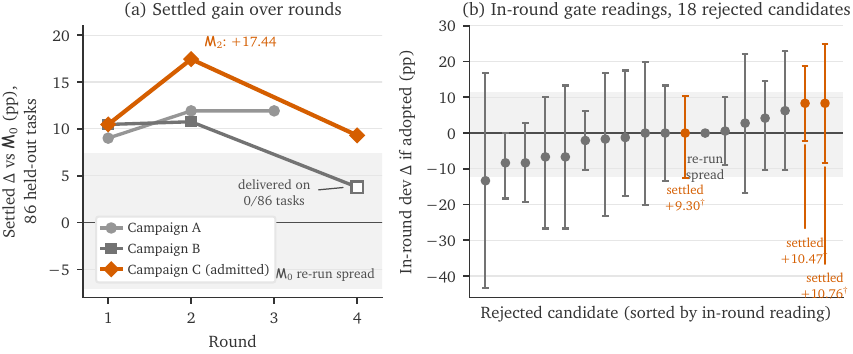}
  \caption{%
    \textbf{Held-out gains are consistent across evolution runs, and finer than the in-round gate
    can resolve.}
    (a) Task success of each final package relative to $\mathcal{M}_0$ on the 86 held-out
    tasks: Run A plateaus, Run B's round-4 package is never invoked (open marker), and
    Run C rises for two rounds and then reverses. The band is the difference between two
    runs of the identical $\mathcal{M}_0$ package.
    (b) The dev-split reading of each rejected candidate at the moment of rejection.
    Every interval contains zero, and the three marked candidates later cleared zero on
    the held-out split.
  }
  \label{fig:evolution}
\end{figure}

Iteration compounds. Along the lineage Run C admitted, the second round adds $+6.98$ points on top of the first, with an interval excluding zero (\cref{fig:evolution}a). Not every round lands: Run A plateaus, its round-3 candidate scoring exactly where round 2 did, and Run C's fourth round gives most of that gain back.

The gains are execution gains. Required-write recall moves with task success across the nine final packages ($r = 0.97$), and each package's write-recall delta sits within about two points of its success delta, $+17.73$ against $+17.44$ at the peak. The packages work by making the agent issue more of the right write calls, not by exploiting slack in the evaluation.

One reading of the figure has to be ruled out. A round that looks like a diminishing return may instead have produced a package that is never invoked: Run B's round-4 candidate scores inside the $\mathcal{M}_0$ re-run spread, and its trace records the cause, the memory reaching none of the 86 held-out tasks against 18.50 invocations per episode for the round-2 candidate of the same line. That is a broken binding, not a diminishing return. The in-round gate is also finer-grained than it can resolve, which \cref{fig:evolution}b shows and \cref{sec:evolution-stability} takes up.

\subsubsection{Two Meta-Agent Implementations Converge}
\label{sec:metaagent-swap}

If the evidence pool rather than the Meta-Agent is what carries recursive evolution, the Meta-Agent should be replaceable wholesale. We test exactly that. Holding the entire protocol of \cref{sec:evolve} fixed, the evidence pool, the localization procedure, the update set and the gate, we re-implement the Meta-Agent from scratch on a second, independent agent stack: Run A runs the original implementation, built on Claude Code, and Runs B and C run the re-implementation, built on DeepSeek Harness, against the same frozen 86-task split, the same $k=4$ protocol and the same deployment model.

\begin{table}[t]
\centering
\caption{%
  \textbf{Swapping the Meta-Agent from Claude Code to DeepSeek Harness reproduces the
  gain, the endpoint and the learned components.}
  All packages are evaluated on the frozen 86-task split at $k=4$ (\cref{tab:evolution-dynamics});
  $\Delta$ carries a paired task-clustered bootstrap 95\% CI, $p$ is McNemar's exact test,
  and $^{\dagger}$ marks an interval excluding zero. The implementation contrast pairs the
  DeepSeek Harness round-1 package (Run B) against the Claude Code round-2 package
  (Run A); the threshold contrast pairs Run C's final package against the same
  Run B package. The A/A row re-runs the byte-identical $\mathcal{M}_0$ package and
  sets the instrument's resolution.
}
\label{tab:metaagent}
\small
\setlength{\tabcolsep}{6pt}
\begin{tabular}{@{}l ccc@{}}
\toprule
\textbf{Comparison} & \textbf{$\Delta$ (points)} & \textbf{95\% CI} & \textbf{$p$} \\
\midrule
\multicolumn{4}{@{}l}{\itshape Settled gain over $\mathcal{M}_0$} \\
\quad Claude Code (Run A) & \hc{11.92}{$+11.92^{\dagger}$} & $[+4.65,\,+19.19]$ & \na \\
\quad DeepSeek Harness (Run B) & \hc{10.47}{$+10.47^{\dagger}$} & $[+3.78,\,+17.15]$ & 0.0022 \\
\quad DeepSeek Harness, progressive (Run C) & \hc{9.30}{$+9.30^{\dagger}$} & $[+1.45,\,+17.44]$ & 0.0068 \\
\midrule
\multicolumn{4}{@{}l}{\itshape Direct paired contrasts} \\
\quad DeepSeek Harness $-$ Claude Code & $-1.45$ & $[-7.85,\,+4.65]$ & 0.72 \\
\quad Progressive $-$ fixed threshold & $-1.16$ & \na & 0.77 \\
\quad Byte-identical $\mathcal{M}_0$, re-run (A/A) & $+0.00$ & $[-6.98,\,+7.27]$ & \na \\
\bottomrule
\end{tabular}
\end{table}

The two implementations are statistically interchangeable (\cref{tab:metaagent}). Each clears $\mathcal{M}_0$ with an interval excluding zero, $+11.92$ on Claude Code and $+10.47$ and $+9.30$ on DeepSeek Harness, three evolution runs converging on a gain of about ten points. Paired task-by-task, the DeepSeek Harness package sits $-1.45$ points from the Claude Code package, CI $[-7.85, +4.65]$ ($p = 0.72$), inside the $[-6.98, +7.27]$ band that two runs of the byte-identical $\mathcal{M}_0$ package span: swapping the Meta-Agent moves the result by less than the instrument's own noise.

The convergence is mechanistic as well as numerical. Opened up, the champion packages of the two implementations arrive at the same component family: a working-memory field tracking service-request authorization, an execution-gate check and a set of anti-escalation skills. Two Meta-Agents that share no code, prompted and orchestrated independently, distill the same sixteen failures into the same repair. What the loop learns is determined by the evidence it diagnoses from, not by the machinery that carries the diagnosis.

\subsubsection{Gated Updates Preserve Existing Abilities}
\label{sec:evolution-stability}
%
%
%

Gated evolution is safe to iterate: accepted updates preserve what the memory already supports. They broke 4 of the 42 dev tasks the working base already solved at every attempt ($9.5\%$), against $25.9\%$ for re-running a byte-identical package on the same tasks (exact binomial $p = 0.013$), and none of the four dropped to zero. Rejected candidates broke them at a rate indistinguishable from that null: regression on anchor tasks is dominated by decode noise rather than by the patch.

What the gate discards is noise, not demonstrated gains. Two runs of an unchanged package on a 10--14 task dev split differ by $[-12.1, +11.1]$ points, every one of the 18 rejected candidates has a dev interval containing zero, and 17 of the 18 sit inside their own size-matched band (\cref{fig:evolution}b).

The loop stays safe even without the dev gate, over the horizon we can test. A fourth evolution run admits every candidate that passes the repair screen and records the dev verdict without enforcing it. The dev gate condemned two of its three patches in round, at $-10.4$ and $-6.2$ points, but on the frozen split that harm never appears: all three packages land between $+14.5$ and $+18.0$ over $\mathcal{M}_0$ with every interval excluding zero, and both negative readings sit inside the unchanged-package band above. The gate errs on the side of caution rather than harming the memory.

The memory only grows, and it can afford to. Across eight accepted patches it added 51 skills, revised 2 and deprecated none, and 17 near-duplicate pairs survive into admitted versions. Yet no single skill carries the gain: rebuilding one accepted package without its five procedure skills, then without the two that remained, then without its largest skill alone, moves the held-out result by $-2.3$ to $+1.7$ with every interval containing zero, and all three variants still clear $\mathcal{M}_0$. The evolved memory is redundant rather than fragile, which is what makes a pruning operator a natural addition to the update set rather than a risk.

\subsubsection{Evolved Skill Memory Transfers across Tasks}
\label{sec:transfer}

Evolved memory transfers, and it transfers where its repairs still apply. The memory is distilled from failures on sixteen training tasks, and the split was fixed before any evolution run ran: sixteen tasks the Meta-Agent reads, twelve the gate screens on, and eighty-six that nothing touches until the held-out evaluation. Nine packages from three evolution runs are evaluated on those eighty-six. Every package that reaches them clears $\mathcal{M}_0$ by between $+9.01$ and $+17.44$ points with an interval excluding zero (\cref{tab:evolution-dynamics}).

Invocation on held-out tasks is not automatic, and the structured trace records how often it occurs. On held-out tasks the evolved memory is invoked on 78 to 86 of the 86, between 2.45 and 18.50 times per episode depending on the package. $\mathcal{M}_0$ holds no skills and is invoked on none, so the counter tracks the memory rather than the harness. The single package that is invoked on zero held-out tasks is also the only one that fails to clear $\mathcal{M}_0$. The transferred gain is therefore not a generic prompting effect: the skills are retrieved and applied to orders, customers and policies other than those they were written from.

Where the held-out pool retains no failures of the kind the memory repairs, there is nothing left to transfer. Two $\tau^2$-Airline lineages are evaluated on the tasks their own runs never saw, 25 and 29 of them, and neither result separates from zero. Both held-out sets start high, at $73.0\%$ and $81.0\%$: the evolution run records showed $\tau^2$-Airline had no unused improvable tasks left, the failures worth repairing having already been allocated to train and dev. The airline packages also reach less, 15 of 25 and 13 of 29 tasks, so the memory is both less needed and less often invoked.

We therefore state transfer narrowly. Memory evolved from sixteen failures raises success on eighty-six tasks it never saw, by nine to seventeen points, and it does so across two Meta-Agent implementations (\cref{sec:metaagent-swap}) and three admission-threshold settings. It does so where the held-out tasks still contain the kind of failure the memory repairs, and not where they do not.

\subsubsection{Evolved Skill Memory Transfers across Models}
\label{sec:transfer-models}

The preceding subsection held the model fixed and varied the tasks; here we hold the tasks fixed and vary the model. This is the question the single-source design of \cref{sec:setup} was built to answer: a memory evolved once, on one model, is shipped unchanged to models that took no part in producing it, so whatever it earns there is earned by a memory that was never fitted to them. On $\tau^2$-Retail it adds $17.8$ points to GPT-5.6 Sol, $15.6$ to Claude Opus 5 and $4.8$ to Gemini 3.7 Flash, the first two with intervals excluding zero (\cref{tab:frontier}).

It is not a crutch for weaker models. The same package lifts every model it is given to, including the deployment model it was evolved on by $23.3$ points, and the strongest target model also ends highest on that benchmark, at $87.9$ against the $81.4$ the package reaches on the model it came from. A package that merely compensated for missing capability would have the least room left exactly where capability is highest.

What the package carries decides where it transfers, and the receiving model does not. On SkillFlow transfer broadly follows scale, from no gain on the smallest model to $16.6$ and $13.5$ points on the two largest open-weight ones; on $\tau^2$-Retail it does not, the smallest model gaining $13.4$ points with an interval excluding zero while three larger ones show no measurable gain. A SkillFlow package carries procedure, and any model able to run the procedure can use it. A $\tau^2$ package carries discipline, what to verify and when a goal is still open, and discipline is worth only what the target model's own failures make it worth. Claude Opus 5 makes the point with the model held fixed: $15.6$ points on $\tau^2$-Retail, and one point on $\tau^2$-Airline where its own baseline already sits near ninety. Remaining headroom is therefore a ceiling on what a memory can buy, not a prediction of it, and two nearly identical starting levels show it: at $72.4$ and $73.5$ the same package is worth $+15.6$ and $+4.8$.

\begin{table}[t]
\centering
\caption{%
  \textbf{One package, evolved on a mid-sized deployment model, lifts frontier models it
  never saw.} Each configuration runs the full task set at $k=4$ with zero null episodes;
  $\Delta$ carries a paired task-clustered bootstrap 95\% CI and $^{\dagger}$ marks an
  interval excluding zero. Rows are ordered by the level the agent reaches alone.}
\label{tab:frontier}
\small
\setlength{\tabcolsep}{5pt}
\begin{tabular}{@{}llccc@{}}
\toprule
\textbf{Model} & \textbf{Domain} & \textbf{agent alone} & \textbf{$+$ \ours{}} & \textbf{$\Delta$ (95\% CI)} \\
\midrule
GPT-5.6 Sol   & $\tau^2$-Retail  & 58.33 & \best{76.10} & \hc{17.76}{$\mathbf{+17.76}^{\dagger}$} \tiny{$[+11.84,+23.90]$} \\
Claude Opus 5 & $\tau^2$-Retail  & 72.37 & \best{87.94} & \hc{15.57}{$\mathbf{+15.57}^{\dagger}$} \tiny{$[+10.96,+20.39]$} \\
Gemini 3.7 Flash & $\tau^2$-Retail & 73.46 & \best{78.29} & \hc{4.82}{$+4.82$} \tiny{$[-0.22,+10.09]$} \\
GPT-5.6 Sol   & $\tau^2$-Airline & 79.00 & \best{86.00} & \hc{7.00}{$\mathbf{+7.00}^{\dagger}$} \tiny{$[+1.50,+13.00]$} \\
Claude Opus 5 & $\tau^2$-Airline & 89.50 & \best{90.50} & \hc{1.00}{$+1.00$} \tiny{$[-3.00,+4.00]$} \\
Gemini 3.7 Flash & $\tau^2$-Airline & \best{86.50} & 85.00 & \textcolor{black!55}{$-1.50$} \tiny{$[-5.50,+2.00]$} \\
\bottomrule
\end{tabular}
\end{table}

\subsubsection{Case Study: Targeted Skill Evolution}
\begin{figure}[H]
  \centering
  \includegraphics[width=\linewidth]{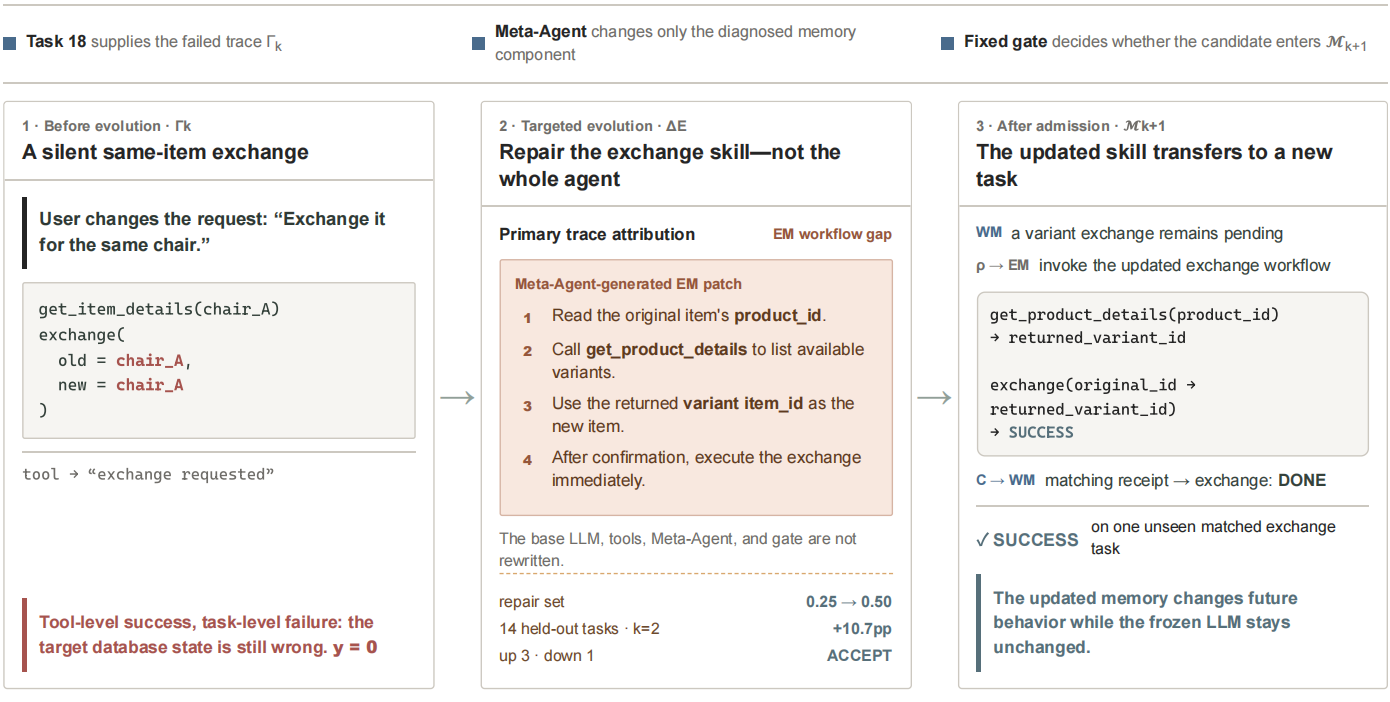}
  \caption{
    \textbf{Illustration of component-specific Skill Memory evolution.}
  }
  \label{fig:case-evolution}
\end{figure}

As shown in \cref{fig:case-evolution}, the structured trace exposes a same-item exchange error that can be assigned to the experiential-memory workflow; the accepted patch changes that workflow rather than rewriting the entire memory. A failed exchange trace reuses the original item identifier as the replacement, yielding a tool acknowledgment but an incorrect task state (left). The Meta-Agent attributes this to an experiential-memory gap and patches that component to retrieve a valid variant identifier before the exchange (middle); once admitted, the updated memory succeeds on an unseen matched task (right). The $0\!\rightarrow\!100\%$ refers to that single task.

%
%

\newcommand{\ttaN}{87}
\newcommand{\ttaBudget}{4}
\newcommand{\ttaBase}{34.5}        
\newcommand{\ttaSeedOne}{32.2}     
\newcommand{\ttaSeedFour}{58.6}    
\newcommand{\ttaOurs}{60.9}        
\newcommand{\ttaRetryGain}{26.4}   
\newcommand{\ttaLearnGain}{2.3}    

\subsection{Test-Time Adaptation on Isolated Tasks}
\label{sec:tta}

\Cref{sec:evolve} builds one memory per domain and asks it to generalise. The same machine also runs in a second mode, \emph{test-time adaptation}: the memory is rebuilt for a single task, from that task's own failed trajectory, and is applied to that same task on the next attempt. This is the regime for isolated tasks that share no tools or policies with any other task, where cross-task evolution has nothing to carry (\cref{sec:overall}).

\hpara{Setup.}
We evaluate on Terminal-Bench 2.1~\citep{terminalbench2}, \ttaN{} terminal tasks executed in their official Docker environments with the Terminus-2 agent and \texttt{doubao-seed-2-0-pro} as the policy model. Every configuration gets a budget of \ttaBudget{} attempts per task, stops at the first success, and counts the task failed if the budget runs out; configurations share the agent, the task images, the reasoning effort, and the interaction budget. Between attempts the Meta-Agent receives the task instruction, the failed trajectory, and one bit stating that a hidden verifier scored the attempt zero. It never sees the verifier, the tests, or any expected output, and the restriction is enforced in the harness rather than in the prompt. The adaptation and no-adaptation configurations share their first rollout verbatim, so their contrast is paired and carries none of the first attempt's variance.

\begin{table}[t]
\centering
\caption{%
  \textbf{Terminal-Bench 2.1, decomposed: the attempt budget carries the headline, and the
  memory terms are read at a matched budget.}
  Every configuration stops at its first success. \emph{Solved} counts tasks solved within
  budget at one rollout per attempt. $\Delta$ and $p$ compare each row with the row above it
  on the same tasks, $p$ from McNemar's exact test; the last column names what that
  comparison isolates.}
\label{tab:tta}
\small
\setlength{\tabcolsep}{5pt}
\begin{tabular}{@{}l c c c c l@{}}
\toprule
\textbf{Configuration} & \textbf{Budget} & \textbf{Solved} & \textbf{$\Delta$} & \textbf{$p$} & \textbf{$\Delta$ isolates} \\
\midrule
Terminus-2 (baseline) & 1 & 30/87 \tiny{(34.5)} & \na & \na & \na \\
\quad $+$ seed memory & 1 & 28/87 \tiny{(32.2)} & $-2.3$ & 0.824 & the layer alone \\
\quad $+$ seed memory, retry & 4 & 51/87 \tiny{(58.6)} & $+26.4$ & $<10^{-4}$ & the attempt budget \\
\rowcolor[gray]{.9}
\quad $+$ test-time adaptation & 4 & \best{53/87} \tiny{(60.9)} & $+2.3$ & 0.774 & learning, matched budget \\
\bottomrule
\end{tabular}
\end{table}

\hpara{Results.}
\Cref{tab:tta} reports the four configurations, and its last column names what each step isolates. Test-time adaptation solves 53 of the \ttaN{} tasks, $\ttaOurs{}\%$, the best of the four and $+26.4$ points over the single-attempt baseline. That headline is not a learning effect, and the table says so: retrying the seed memory is worth the same $+\ttaRetryGain{}$ points on its own, 23 tasks flipping to solved and none flipping back ($p<10^{-4}$), while at a matched budget of four attempts adaptation adds $+\ttaLearnGain{}$ points, 7 tasks won against 5 lost ($p=0.774$). The memory-control layer without learning is likewise inside noise, at $-2.3$ points ($p=0.824$). On this metric the attempt budget therefore explains the headline and both memory terms sit within run-to-run variation. Seven tasks, among them \texttt{compile-compcert} and \texttt{qemu-alpine-ssh}, are nevertheless solved only under adaptation and by no other configuration.

\hpara{The per-attempt view.}
\Cref{tab:main} scores this benchmark by \texttt{avg@4}, like every other column; \cref{tab:tta} instead counts tasks solved within budget, which is the metric the benchmark's own protocol reports and the one the decomposition above needs. Solved-within-budget is, however, the wrong instrument for a learning effect, because extra attempts move it: over a quarter of the tasks sit in a regime where the agent can solve them but not reliably, and two runs of the identical seed memory differ by 5.7 points with nothing changed. We therefore also measure \emph{per-attempt} success at four untruncated rollouts, a quantity no retry budget can buy (\cref{tab:tta-peratt}). On the 56 tasks whose adapted memory carries a learned skill, avg@4 rises from $17.4\%$ to $21.9\%$, $+4.5$ points, with 14 tasks improving and 7 degrading. What we read is not that number but its consistency: all four cuts of these runs, two metrics on two task sets, move the same way, by $+2.3$ to $+4.5$ points. Every interval still contains zero at this sample size, so we report a direction rather than an effect. On the one task we examined at depth the effect is decisive: at 16 rollouts per configuration the bare agent solves it 8 times, the seed memory 7, and the adapted memory 15 ($p=0.006$). Invocation is verified rather than assumed: kernel counters show the learned skill reaching the agent in 53 of 57 second-round trials, and a post-hoc audit flagged none of the 93 learned skills as containing a task-specific value. Within-task adaptation therefore complements the cross-task mode of \cref{sec:evolve}: its gains concentrate on individual hard tasks and on per-attempt reliability, the margins a retry budget cannot buy.

\begin{table}[t]
\centering
\caption{%
  \textbf{Per-attempt success, the metric extra attempts cannot move.}
  Four untruncated rollouts per configuration, seed memory against the memory test-time
  adaptation produced. \emph{Learned} is the 56 tasks whose adapted memory carries a learned
  skill; \emph{all} splices those with the remaining tasks, which adaptation leaves
  unchanged. CIs are paired task-clustered bootstrap $95\%$ intervals.}
\label{tab:tta-peratt}
\small
\setlength{\tabcolsep}{6pt}
\begin{tabular}{@{}l c c c c c@{}}
\toprule
\textbf{Metric} & \textbf{Tasks} & \textbf{seed} & \textbf{adapted} & \textbf{$\Delta$} & \textbf{95\% CI} \\
\midrule
avg@4, learned  & 56 & 17.4 & \best{21.9} & \hc{4.5}{$+4.5$} & \tiny{$[-0.9,\,+9.8]$} \\
pass@4, learned & 56 & 39.3 & \best{42.9} & \hc{3.6}{$+3.6$} & \tiny{$[-8.9,\,+16.1]$} \\
avg@4, all      & 87 & 46.1 & \best{48.9} & \hc{2.9}{$+2.9$} & \tiny{$[-0.6,\,+6.3]$} \\
pass@4, all     & 87 & 60.9 & \best{63.2} & \hc{2.3}{$+2.3$} & \tiny{$[-5.7,\,+10.3]$} \\
\bottomrule
\end{tabular}
\end{table}

\section{Related Work}
\label{sec:related}
\subsection{Experiential Memory and Agent Skills}
Augmenting LLM agents with reusable experiential memory and agent skills has shown clear benefits on challenging tasks. Existing methods store experience in different forms. Voyager \citep{wang2023voyager} stores executable code skills, AWM \citep{wang2024awm} induces natural-language workflows, ExpeL \citep{zhao2024expel} extracts insights from successes and failures, Buffer of Thoughts \citep{yang2024bot} and ReasonFlux \citep{yang2025reasonflux} maintain thought templates distilled from previous solutions, and Dynamic Cheatsheet \citep{suzgun2026dynamic} curates notes at test time. More recently, Agent Skills \citep{anthropic2025skills} provides a standardized format for representing these different forms of reusable experience. Subsequent work such as SkillOpt \citep{yang2026skillopt} treats skill files as trainable parameters and optimizes them based on evaluation feedback, while SkillComposer \citep{zhang2026skillcomposer} learns to create, improve, and merge skills.

Despite this progress, existing work primarily focuses on what to store, with less attention paid to when and how stored skills should be invoked during execution. In many systems, skills are either injected into the context as a fixed block or the agent is left to decide on its own when to invoke a skill and which one to use. A recent benchmark shows that agents often fail to select the appropriate skill on their own \citep{liu2026skillswild}, while injecting inaccurate skills can even hurt task completion \citep{li2026skillsinjector}. Some systems do condition retrieval on execution context: AutoGuide \citep{fu2024autoguide} stores guidelines with an explicit condition on the situation in which they apply, and SGDR \citep{li2026onlineskill} retrieves sub-procedures at every step by matching both the task goal and the current page state. In both cases the conditioning signal is the raw observation, matched by similarity, so retrieval follows what the environment currently looks like rather than what the agent has confirmed it has achieved. \ours{} instead couples experiential memory with working memory, grounding when and which skills to invoke in the agent's current task state. This enables more precise skill invocation while keeping irrelevant knowledge out of the context.

\subsection{Working Memory for LLM Agents}
Working memory tracks the current task state and supports long-horizon decision making \citep{sumers2024coala}. Since the raw interaction history quickly becomes long and noisy, existing works maintain this state in more explicit forms. StateAct \citep{rozanov2025stateact} asks the model to write and update a structured state at every step, while ReflAct \citep{kim2025reflact} has it reflect on the current state before acting. Magentic-One \citep{fourney2024magenticone} maintains a task ledger of known facts and a progress ledger of open goals, while StateFlow \citep{wu2024stateflow} constrains execution with a hand-written state machine. StructAgent \citep{wu2026structagent} keeps a compact representation of task progress and admits a progress update only through a verifier-backed state transition.

However, an explicit state is not necessarily a trustworthy state. In existing systems, state updates are often either prescribed by fixed rules or written by the model itself, with limited grounding in what the tools actually returned. Even when a separate LLM validator is introduced \citep{chang2025sagallm}, using another model for validation does not fundamentally resolve the reliability problem. Handwritten rules generalize poorly when a task deviates from the designed flow, while working states maintained by the agent itself remain vulnerable to omissions and hallucinations, causing the agent to lose track of completed actions or record progress that was never actually made \citep{huang2025latentstate}. Where verification is present \citep{wu2026structagent}, the verified state is consumed by the agent's own next action, and no procedural memory is retrieved from it. \ours{} addresses this limitation by separating state proposal from state commitment. Its kernel updates progress entries only when they are supported by actual tool results, so both action decisions and skill invocation are grounded in the agent's verified task state.

\subsection{Recursive Self-Improvement of LLM Agents}
\label{sec:rsi}
Recursive self-improvement has a long formal history: a G\"odel machine \citep{schmidhuber2003godel} rewrites its own code once it can prove the rewrite improves future utility. With language models the idea became practical at the scaffolding level, where an improver program is applied to itself \citep{zelikman2024stop, fernando2024promptbreeder}. Recent agent systems differ mainly in which layer they rewrite: their own runtime logic or codebase \citep{yin2025godelagent, robeyns2025sica}, the workflow graph or module composition \citep{zhang2025aflow, shang2025agentsquare}, the tool set \citep{qiu2025alita}, or the weights \citep{hu2025adas, zhang2025dgm, zweiger2025seal, zhai2025agentevolver, simonds2025ladder}. AlphaEvolve \citep{novikov2025alphaevolve} narrows the target to the artifact rather than the agent, holding its own search procedure fixed; MetaSkill-Evolve \citep{wang2026metaskill} goes the other way, evolving on a slow timescale the meta-skill that drives improvement on the fast one.

The narrowest layer, and the one \ours{} occupies, is the memory itself: experience is written back after each task and maintained through add, update and delete operations \citep{chhikara2025mem0, xu2025amem}, distilled into reusable strategies \citep{ouyang2026reasoningbank}, or repaired from failed runs \citep{zhu2025agentdebug, zhang2026memskill}. Memento \citep{zhou2025memento} freezes the base model and adapts only a case bank; EvolveMem \citep{liu2026evolvemem} evolves the memory architecture and accepts a change when benchmark performance improves; a plug-in controller gates which updates are consistent enough to apply at all \citep{chen2026prologue}.

Although these methods differ in what they rewrite, they agree on how a rewrite is admitted: by the model's own judgment of usefulness \citep{shinn2023reflexion}, or by an improvement in a single aggregate number \citep{khattab2024dspy, agrawal2026gepa, yang2026skillopt, liu2026evolvemem}. Stricter gates exist, from novelty checks to paired significance tests and consistency verification \citep{wang2026sage, shawn2026pace, lam2026ssgm}, but they decide whether an update is kept rather than why the agent failed. Such a signal cannot say which part of the agent produced the failing runs, so the unit diagnosed and the unit patched are rarely the same. Attribution is also hard to add after the fact, reaching $53.5\%$ for the responsible agent and $14.2\%$ for the decisive step over unstructured logs \citep{zhang2025which}, while a wider modifiable surface has been observed to degrade behaviour \citep{shao2026misevolve}. \ours{} confines recursion to the memory-control layer, holding the base model and the improvement procedure fixed. Because every step records the working state, the invoked skill, the action and the observation, the Meta-Agent attributes each diagnosed failure to one of the four memory components and patches only those implicated, behind a gate that requires the diagnosed tasks to be repaired without regressing a held-out split.

\section{Conclusion}
\label{sec:conclusion}

We present \ours{}, a recursive Experiential--Working Memory architecture that reframes recursive self-improvement as an operation on an externalized memory-control layer rather than on the model or the agent. Working Memory maintains a verified task state, that state grounds skill invocation in Experiential Memory, and the coupling emits a structured trace linking task states, invoked skills, actions and outcomes. A fixed Meta-Agent reads that trace, attributes each diagnosed failure to a memory component, and patches only the components implicated, behind a fixed validation gate; the base model and the improvement procedure never change. Across four long-horizon benchmarks and ten target models, \ours{} improves task success in 35 of the 37 completed model--benchmark pairs, adds $+17.8$ and $+15.6$ points to GPT-5.6 Sol and Claude Opus 5 on $\tau^2$-Retail with a memory evolved on neither, leads the base agent in every horizon quartile by $+17.0$ to $+44.7$ points, and localizes an injected fault at $64.8\%$ against $13.0\%$ from the task outcome alone. These results suggest that the memory-control layer, rather than the weights, can serve as the trainable surface of a frozen agent, yielding improvement that is attributable to a component, reversible, and portable across models.

\bibliography{references}

\newpage
\appendix

\section{Statistical Protocol}
\label{app:stats}

Reported intervals are paired task-clustered bootstrap 95\% confidence intervals over $10{,}000$ resamples of tasks. We resample tasks rather than episodes because the $k$ attempts at one task share its goal, tools and environment; pairing is on the task, so each resample compares two variants on the same tasks. We mark intervals that exclude zero and do not call a difference an effect when its interval includes zero.

Interval width is not the only source of uncertainty. Re-running one $\tau^2$-Retail package unchanged, three days apart on the same tasks, moves task success by $+0.00$ points with an interval of $[-6.98, +7.27]$, so on that domain we treat differences of a few points as within run-to-run variation regardless of their interval. For the same reason each analysis draws its own evaluation population: absolute levels are comparable within a table or a panel, not across them. Where a contrast is between two binary outcomes on the same tasks we report McNemar's exact two-sided test alongside the interval.

\section{Harness Ablation: Where the Gain Comes From}
\label{app:harness-ablation}

A memory-control layer is itself a piece of engineering, and a reader is entitled to ask how much of our reported gain is bought by that engineering rather than by anything the loop learns. We answer it by ablating the harness: run the benchmark's own agent alone, then run the same agent inside \ours{} carrying $\mathcal{M}_0$, the neutral starting package that holds no learned content, at the same harness version, the same split and the same $k$. The difference between those two configurations is everything our memory-control layer contributes before a single patch is admitted.

\begin{table}[h]
\centering
\caption{%
  \textbf{The harness on its own contributes nothing measurable.} Bare agent against
  $\mathcal{M}_0$ on $\tau^2$-Retail, 86--88 tasks per model at $k=4$, zero null episodes,
  one harness commit and byte-identical decoding settings. Every interval contains zero and
  the point estimates split two positive, two negative.
}
\label{tab:machine-tax}
\small
\begin{tabular}{@{}lccccc@{}}
\toprule
\textbf{Target model} & \textbf{bare} & \textbf{$\mathcal{M}_0$} & \textbf{$\Delta$} & \textbf{95\% CI} & \textbf{up/down} \\
\midrule
GPT-OSS-20B   & 45.35 & 50.58 & $+5.23$ & $[-1.16,\,+11.63]$ & 31 / 20 \\
Qwen3.5-9B    & 77.84 & 77.56 & $-0.28$ & $[-5.11,\,+4.83]$  & 23 / 29 \\
Qwen3.6-35B   & 78.78 & 78.20 & $-0.58$ & $[-5.81,\,+4.65]$  & 21 / 25 \\
GLM-4.7-Flash & 61.05 & 61.92 & $+0.87$ & $[-4.94,\,+6.98]$  & 27 / 26 \\
\bottomrule
\end{tabular}
\end{table}

Every interval contains zero and the point estimates split two positive and two negative, which is the shape of no systematic effect rather than of a small effect we cannot resolve. Decomposing the same four models into a harness term and an evolution term shows where the effect does live. The one model that gains measurably, GPT-OSS-20B, gains it in the evolution term, $+10.17$ with an interval excluding zero, while its harness term stays inside noise; the other three sit inside noise in both terms. Whatever \ours{} is worth, it is worth it through what the loop admits, not through the layer that carries it.

\begin{table}[h]
\centering
\caption{%
  \textbf{Where the effect lives.} The harness term is $\mathcal{M}_0$ minus bare, evolution is
  the final package minus $\mathcal{M}_0$, total is the final package minus bare.
  $^{\dagger}$ marks an interval excluding zero.
}
\label{tab:decomposition}
\small
\begin{tabular}{@{}lccc@{}}
\toprule
\textbf{Target model} & \textbf{harness} & \textbf{evolution} & \textbf{total} \\
\midrule
GPT-OSS-20B   & $+5.23$ & $\mathbf{+10.17}^{\dagger}$ \tiny{$[+4.07,+16.28]$} & $\mathbf{+15.41}^{\dagger}$ \tiny{$[+8.14,+22.38]$} \\
Qwen3.5-9B    & $-0.28$ & $+1.99$ & $+1.70$ \\
Qwen3.6-35B   & $-0.58$ & $+0.29$ & $-0.29$ \\
GLM-4.7-Flash & $+0.87$ & $+0.00$ & $+0.87$ \\
\bottomrule
\end{tabular}
\end{table}

\section{Benchmarks and Splits}
\label{app:splits}
\hpara{SkillFlow.}
SkillFlow~\citep{skillflow} is a published benchmark for lifelong skill discovery and evolution, not one we authored. Its 166 tasks are organised into 20 families of eight to nine tasks, and within a family the tasks are built to follow one domain-agnostic execution flow, so a repair earned on one task of a family is expected to apply to its siblings.

\begin{multicols}{2}
\begin{itemize}[topsep=2pt,itemsep=0pt,parsep=0pt,leftmargin=1.3em]
  \item \small Compensation-Scenario-Modeling
  \item \small Cross-Format-Data-Reconciliation
  \item \small DMAIC-Quality-Analysis
  \item \small Distribution-Center-Auditing
  \item \small Document-Fraud-Detection
  \item \small Embedded-Data-Repair
  \item \small Financial-Statement-Rolling
  \item \small HWPX-Document-Automation
  \item \small Healthcare-Cost-Benefit-Analysis
  \item \small Industry-Correlation-Analysis
  \item \small Inventory-and-Finance-Integration
  \item \small Medical-Data-Standardization
  \item \small OCR-Data-Extraction
  \item \small Operational-Recovery-Planning
  \item \small PPT-Formatting-Optimization
  \item \small Production-Capacity-Planning
  \item \small SEC-13F-Financial-Analysis
  \item \small Sales-Pivot-Analysis
  \item \small Supply-Chain-Replenishment
  \item \small Weighted-Risk-Assessment
\end{itemize}
\end{multicols}
 Two task identifiers appear in two families each, so a name-deduplicated count gives 164; we report the 166-pair count throughout.

Scoring is programmatic. Each task carries its own verifier script and returns a binary reward, so no model judge enters the metric and no judge-agreement question arises. The absolute levels are low because the tasks demand a strict execution procedure rather than because the verifier is noisy: under the same verifier the deployment model moves from $34.6\%$ to $51.4\%$ and Qwen3.6-27B from $42.2\%$ to $58.7\%$, which a scoring artefact could not produce.

\hpara{How a package is evolved on $\tau^2$-Bench.}
Before any run, the tasks of a domain are partitioned into three disjoint splits and the partition is frozen. The loop reads failed episodes only from the \emph{evolve} split, so every patch is written against a failure the Meta-Agent has actually seen. The \emph{dev} split is the gate's evidence and nothing else: it deliberately contains anchor tasks the working memory already solves, so a candidate that repairs its target while breaking something else shows up as a regression rather than as a net gain. The \emph{test} split is touched only at the held-out evaluation, when every package in a lineage is re-evaluated on it at $k=4$ under one harness version, which is what makes the packages of different evolution runs comparable on one scale.

\hpara{How a package is evolved on SkillFlow.}
Here we follow the benchmark's own protocol rather than imposing ours. The agent starts a family with no skills, works its tasks in sequence, and writes skill patches from its own trajectories and the task rubric; the family, not the individual task, is the unit that carries memory, and a family ends with the template that won inside it. There is therefore no held-out task split on SkillFlow, and the template is in-sample with respect to the tasks it was selected on. What is out of sample is the target model: the package is built once on the deployment model and then shipped unchanged to models that took no part in building it, which is the contrast the main table draws.

\hpara{How adaptation works at test time.}
Terminal-Bench 2.1 has no cross-task structure to evolve against, so the same machinery runs in a second mode described in \cref{sec:tta-method}. The scope of the evidence pool is the only thing that changes: memory is built for one task, from that task's own failed attempt, and applied to that same task on the next attempt. Between attempts the Meta-Agent receives the task instruction, the failed trajectory and one bit stating that a hidden verifier scored the attempt zero, and nothing else; the restriction is enforced by the harness rather than requested in a prompt. Because retrying alone is a strong confound on this benchmark, the comparison is against a budget-matched configuration that retries under the frozen initial memory and shares its first rollout with the adaptation configuration.

\hpara{Split sizes.}
$\tau^2$-Retail uses 16 evolve, 12 dev and 86 test tasks. The two $\tau^2$-Airline lineages use 10 evolve / 15 dev and 11 evolve / 10 dev; in both the split files leave the test field empty and the held-out set is the complement of evolve and dev, 25 and 29 tasks. SkillFlow carries no held-out split: its skills are per-family, one winning template selected within the family it is then used on. SkillFlow results therefore measure transfer across target models, which is what the main table uses them for, and not transfer to unseen tasks.

\section{Compute and Context}
\label{app:compute}

The claim that the gains are not bought with more context or more compute is testable on the four-configuration $\tau^2$-Retail family, where every configuration shares the model, the tasks and the budget and only the memory-control regime differs.

\begin{table}[h]
\centering
\caption{%
  \textbf{More context buys a worse result.} The model-controlled configuration keeps the whole skill
  library standing in context, $3{,}111$ more prompt tokens at the first call than \ours{},
  and is 18 points worse while costing $46\%$ more per success.
}
\label{tab:compute}
\small
\begin{tabular}{@{}lcccc@{}}
\toprule
\textbf{Regime} & \textbf{Success (\%)} & \textbf{First-call prompt} & \textbf{Tokens/episode} & \textbf{Tokens/success} \\
\midrule
Bare agent            & 58.11 & 5163 & 67{,}315 & 115{,}833 \\
Working memory only   & 82.02 & 5636 & 83{,}938 & 102{,}341 \\
Model-controlled invocation & 65.57 & 8274 & 96{,}289 & 146{,}849 \\
\ours{}               & \best{83.55} & 5632 & 84{,}275 & \best{100{,}865} \\
\bottomrule
\end{tabular}
\end{table}

Three quantities are not recoverable from these artefacts and we state them rather than approximate them. The framework does not persist injected skill text into the message log, verified by a byte scan of all four result files, so a per-turn count of rendered memory tokens does not exist; we report the standing-context increment at the first call, the static skill sizes on disk (the retail champion package is 10 skills totalling $10{,}135$ characters) and an off-parity bounce estimate from an earlier run of the same package instead. Dollar cost is unavailable because the cost fields are zero in every simulation, so we report tokens. Terminal-Bench 2.1 and SkillFlow run through container harnesses whose token records have a different shape and are not aggregated here.

\section{Case Studies}
\label{app:cases}

Three of the four cases below come from $\tau^2$-Retail and Terminal-Bench 2.1. SkillFlow supports the evolution case but not the within-task one, and the reason is structural rather than a gap in our records: skills reach a SkillFlow agent as a prompt template injected once, so no working memory, invocation policy or checker exists at run time to observe. This is also why the coupling analysis of \cref{sec:ablation-coupling} runs on $\tau^2$ alone.

\hpara{Component-Scoped Repair, and a Gate That Waits.}
\label{app:case-evolution}

One round of the loop on $\tau^2$-Retail shows the two properties the method claims: a patch scoped to the component the trace blames, and a gate that admits nothing it cannot demonstrate.

\begin{casepanel}{Run A, round 2 \textnormal{\,|\,} $\tau^2$-Retail}
\textbf{What the Meta-Agent decided.} Reading the structured traces of the round's failed episodes, it grouped them into clusters and assigned each to a component \emph{before} proposing anything.

\begin{center}\small
\begin{tabular}{@{}llcl@{}}
\toprule
\textbf{Cluster} & \textbf{Blamed component} & \textbf{Fix} & \textbf{Tasks} \\
\midrule
refund payment method      & experiential memory & add skill  & 13 \\
exchange variant procedure & experiential memory & add skill $\times2$ & 18, 91 \\
modify payment procedure   & experiential memory & add skill  & 40 \\
cancel procedure           & experiential memory & add skill  & 90 \\
modify items procedure     & experiential memory & add skill  & 101 \\
\multicolumn{4}{@{}l}{\itshape \ldots two further experiential-memory clusters} \\
\addlinespace[2pt]
\textcolor{caseaccent}{\textbf{harness}} & \textcolor{caseaccent}{\textbf{the harness itself}} & \textcolor{caseaccent}{\textbf{none}} & 0, 13, 18, 91, 101 \\
\bottomrule
\end{tabular}
\end{center}

Every repaired cluster lands on the experiential memory; not one touches the working-memory spec, the invocation policy or the checkers. The last row is the one worth reading twice. The Meta-Agent attributed a further group of failures to the harness itself and then proposed no fix for them at all. Attribution is not a formality that always ends in a patch: the loop can conclude that a failure is not the memory's to repair, and leave it alone. The round shipped seven skills, one recording an action result, one a piece of domain knowledge and five encoding procedures.

\tcblower
\textbf{What the gate did with it.} It rejected the candidate, and had rejected the round before it.

\begin{artifactbox}{driver.log}
GATE r1: net $-$2.1pp \ CI[$-$10.4, $+$6.2] \ up 2 / dn 3 \ repair .214$\,\to\,$.393 \ \textcolor{caseaccent}{\textbf{REJECT}}\\
GATE r2: net $+$6.2pp \ CI[$-$8.3, $+$22.9] \ up 4 / dn 2 \ repair .214$\,\to\,$.571 \ \textcolor{caseaccent}{\textbf{REJECT}}
\end{artifactbox}

Both candidates raised the repair rate on the tasks they were written against, and both were held back because the dev reading could not be separated from zero: a 12-task split at $k=4$ produced an interval more than thirty points wide. Settled afterwards on the 86 held-out tasks the Meta-Agent never reads, the second of those rejected candidates clears $\mathcal{M}_0$ by $\mathbf{+11.92}$ points with an interval excluding zero (\cref{tab:evolution-dynamics}).
\end{casepanel}

This is the clearest statement of what the gate is and is not. It is not a filter that separates good patches from bad ones; on this evidence budget it cannot be. It is a rule that refuses to commit the working memory to a reading the evidence cannot resolve, and the cost of that conservatism is paid back at the held-out evaluation, where a pre-registered split has the statistical power the in-round dev split lacks.

\hpara{Adaptation Inside a Single Task.}
\label{app:case-tta}

On Terminal-Bench 2.1 the same machinery runs against one task at a time. The task \texttt{mailman} asks the agent to stand up an integrated mail service. Its first attempt failed; the per-round record shows what happened next.

\begin{center}
\begin{tikzpicture}[font=\small, node distance=0pt,
  att/.style={draw=caseink, line width=0.7pt, rounded corners=2pt, inner sep=5pt,
              minimum height=11mm, text width=27mm, align=center},
  crd/.style={draw=coderule, fill=codetint, rounded corners=2pt, inner sep=5pt,
              minimum height=11mm, text width=34mm, align=center, font=\scriptsize},
  fl/.style={-{Latex[length=2.4mm]}, line width=0.8pt, draw=caseink}]
  \node[att] (a1) {\textbf{Attempt 1}\\[1pt] reward \textcolor{caseaccent}{\textbf{0.0}}};
  \node[crd, right=11mm of a1] (skill) {skill written\\[1pt]\texttt{full\_config\_}\\\texttt{validation\_before\_}\\\texttt{completion}};
  \node[att, right=11mm of skill] (a2) {\textbf{Attempt 2}\\[1pt] reward \textbf{1.0}};
  \draw[fl] (a1) -- node[above, font=\scriptsize] {diagnose} (skill);
  \draw[fl] (skill) -- node[above, font=\scriptsize] {invoke} (a2);
\end{tikzpicture}
\end{center}

We report this as an instance of the mechanism, not as evidence that adaptation was necessary: the budget-matched contrast in \cref{sec:tta} puts adaptation $+2.3$ points above pure retrying with an interval that contains zero, and a single task cannot settle what that contrast leaves open.

\begin{casepanel}{The information contract, visible in the artefact}
What the Meta-Agent wrote came from the trajectory and nothing else. Its diagnosis record carries exactly five fields:

\begin{artifactbox}{diagnosis\_r1.json}
root\_cause \ \ \ \ \ \ \ \ The agent made incomplete, unvalidated configuration\\
\phantom{root\_cause \ \ \ \ \ \ \ \ }changes (failed to add required Postfix lines, \ldots)\\[2pt]
\textcolor{caseaccent}{\textbf{evidence\_from\_trajectory}}\\
\phantom{root\_cause \ \ \ \ \ \ \ \ }1. Postfix config edits were truncated; only\\
\phantom{root\_cause \ \ \ \ \ \ \ \ }\ \ \ mydestination was updated, no transport entries\\
\phantom{root\_cause \ \ \ \ \ \ \ \ }2. All list po\ldots\\[2pt]
skill\_id \ \ \ \ \ \ \ \ \ \ \ full\_config\_validation\_before\_completion\\
skill\_title \ \ \ \ \ \ \ \ Validate all config changes and full test pass\\[2pt]
skill\_body \ \ \ \ \ \ \ \ \ For integrated service setup tasks:\\
\phantom{root\_cause \ \ \ \ \ \ \ \ }1. After editing config files, grep to confirm every\\
\phantom{root\_cause \ \ \ \ \ \ \ \ }\ \ \ required setting is present\ldots
\end{artifactbox}

There is no field for the reference solution, the verifier's tests or the gold action sequence, because none of them reaches the Meta-Agent. The evidence field cites only what the failed run itself did.
\end{casepanel}

Five further tasks show the same one-round-to-success shape, among them \texttt{bn-fit-modify}, whose skill forbids the specific shortcut the first attempt took, fabricating a dependency graph instead of fitting one.

\hpara{Procedure Is What the Package Carries.}
\label{app:case-skillflow}

SkillFlow runs no evolution loop of ours, so it cannot illustrate localization. It isolates something else cleanly: how much of a task family's difficulty is procedural. Within a family every task follows one execution flow, and the memory a family ends with is a single template. Replacing that template changes nothing about the model, the tasks or the budget.

\begin{table}[h]
\centering
\caption{%
  \textbf{The same model, the same tasks, a different procedural description.} Tasks solved
  in a family under the canonical template and under the template selected inside that
  family. Selection is in-sample, so these figures measure what procedure is worth, not
  generalization.
}
\label{tab:skillflow-templates}
\small
\begin{tabular}{@{}lcc@{}}
\toprule
\textbf{Family} & \textbf{canonical} & \textbf{selected in family} \\
\midrule
weighted-risk-assessment         & 0/8 & \best{7/8} \\
embedded-data-repair             & 4/8 & \best{8/8} \\
healthcare-cost-benefit-analysis & 1/9 & \best{$\sim$6/9} \\
\bottomrule
\end{tabular}
\end{table}

Six of the twenty families were overridden this way. On \texttt{weighted-risk-assessment} the canonical template solves none of the family's eight tasks and the selected one solves seven: the same model, given the same tasks with a different procedural description, goes from failing all of them to solving nearly all. That is the sharpest available statement of what a procedural package is worth. What generalizes on SkillFlow, and what \cref{tab:main} reports, is a different thing: the transfer of a package to target models that took no part in building it.

\hpara{One Step of EM--WM Coupling.}
\label{app:case-coupling}

Within a task, the property that matters is that working memory changes only on verified evidence. An instrumented $\tau^2$-Retail run records both the per-turn working-memory snapshots and the mechanism events, and the two together catch the moment the guarantee binds.

\begin{casepanel}{A claim the checker refused \textnormal{\,|\,} $\tau^2$-Retail, one episode}
The agent had reached agreement with the simulated user and was about to report the request as handled. One goal in working memory, however, had never been closed by a tool result.

\begin{artifactbox}{mechanism event stream}
kind \ \ \ \ \ truth\_bounce\\
checker \ \ truth\_V3\\
reason \ \ \ \textcolor{caseaccent}{1 executable request(s) are NOT EXECUTED but the}\\
\phantom{reason \ \ \ }\textcolor{caseaccent}{draft does not say so\ldots}\\[2pt]
counters \ ledger\_update\_ok 2 \ \ deliver\_inject 2 \ \ truth\_bounce 1\\
\phantom{counters \ }confirm\_verified 1 \ \ write\_review\_bounce 1
\end{artifactbox}

The agent does not get to mark the goal done by asserting it. The state advances when the environment says so and not before, which is the whole content of the coupling claim.
\end{casepanel}

The counterfactual is available at parity rather than by comparison with a bare agent. The same 114 tasks were run with the checkers removed and everything else held fixed, so the contrast isolates that one component instead of the whole architecture; this is the family of fault configurations \cref{sec:ablation-coupling} reports.

\end{document}